\documentclass[journal]{IEEEtran}
\usepackage{graphicx}
\usepackage{subfigure}
\usepackage{url}
\usepackage{times}
\usepackage{amssymb}
\usepackage{array}
\usepackage[ruled,lined]{algorithm2e}
\usepackage{amsmath}
\usepackage{booktabs}
\usepackage{multirow}
\usepackage{booktabs}
\usepackage{threeparttable}
\usepackage{cite}
\usepackage{enumerate}
\usepackage{rotating}
\newcommand{\minitab}[2][c]{\begin{tabular}{#1}#2\end{tabular}}

\begin{document}

\title{The General Pair-based Weighting Loss \\ for Deep Metric Learning}

\author{Haijun Liu, Jian Cheng, Wen Wang and Yanzhou Su

\thanks{H. Liu, J. Cheng, W. Wang and Y. Su are with the School of Information and Communication Engineering, University of Electronic Science and Technology of China, Chengdu, Sichuan, China, 611731.(Corresponding author: chengjian@uestc.edu.cn)}
}% <-this % stops a space

\markboth{}%IEEE Transactions on Neural Networks and Learning Systems}
{Liu \MakeLowercase{\textit{et al.}}: }

\maketitle

\begin{abstract}
 Deep metric learning aims at learning the distance metric between pair of samples, through the deep neural networks to extract the semantic feature embeddings where similar samples are close to each other while dissimilar samples are farther apart. A large amount of loss functions based on pair distances have been presented in the literature for guiding the training of deep metric learning. In this paper, we unify them in a general pair-based weighting loss function, where the minimize objective loss is just the distances weighting of informative pairs. The general pair-based weighting loss includes two main aspects, (1) samples mining and (2) pairs weighting.  Samples mining aims at selecting the informative positive and negative pair sets to exploit the structured relationship of samples in a mini-batch and also reduce the number of non-trivial pairs. Pair weighting aims at assigning different weights for different pairs according to the pair distances for discriminatively training the network. We detailedly review those existing pair-based losses inline with our general loss function, and explore some possible methods from the perspective of samples mining and pairs weighting. Finally, extensive experiments on three image retrieval datasets show that our general pair-based weighting loss obtains new state-of-the-art performance, demonstrating the effectiveness of the pair-based samples mining and pairs weighting for deep metric learning.
\end{abstract}


\section{Introduction}
\label{sec:intro}
Measuring the distances (similarities) between pair of samples is a core foundation for learning (including detection, tracking and recognition, etc.) in computer community. Being able to first measure how similar a given pair of samples are, the following learning tasks will be a lot simpler. For example, given such distance measure, recognition tasks are simply reduced to the nearest neighbor searching problem. From the perspective of distance measure, metric learning \cite{weinberger2006distance,hu2014discriminative,liu2017status,wang2019ranked,wang2019multi,dong2019mining} and dimensionality reduction \cite{roweis2000nonlinear,balasubramanian2002isomap,sugiyama2007dimensionality,cheng2014silhouette,cheng2015silhouette} methods are with the same objective, all aiming at learning semantic distance metric and embeddings such that similar samples are pulled as close as possible while dissimilar samples are pushed apart from each other.
Metric learning has been applied to a variety of applications, such as image retrieval \cite{wang2019ranked,wang2019multi}, clustering \cite{law2017deep,liu2018sequential}, face recognition \cite{wen2016discriminative}, action recognition \cite{cheng2015silhouette}, person re-identification \cite{zhao2017beyond,yu2018hard}.

Traditional metric learning methods \cite{he2004locality,weinberger2006distance,sugiyama2007dimensionality} focus on learning a Mahalanobis distance metric, which always can be equivalent to the Euclidean distance after some linear projections. Recently, due to the remarkable success of convolutional neural networks \cite{ioffe2015batch,he2016deep}, deep metric learning (DML) methods have attracted a lot of attention. DML discriminatively trains the neural networks guiding by the loss function to directly learn the non-linear projection from the original image space to the semantic feature embedding space. The non-linear projection is implemented by the neural networks. Compared to traditional metric learning methods, which always work on the pre-learned feature representations to learn semantic embeddings, the great advantage of DML is that the neural network can jointly learn the feature representations and semantic embeddings. In other words, DML unifies the feature representations and semantic embeddings into one, directly learning the semantic embeddings.

Addition to the neural network architecture, the loss function plays the most important role in a successful deep neural network framework. Generally, the losses can be classified into two categories, softmax-based losses \cite{ranjan2017l2,wang2017normface,wang2018additive,wang2018cosface,deng2019arcface} and the pair distance-based losses \cite{hadsell2006dimensionality,schroff2015facenet,ohS2016deep,sohn2016improved,wang2019multi,wang2019ranked}. The softmax-based loss aims at solving the classification tasks, predicting the correct labels for every sample. However, the pair distance-based loss aims at learning the distance metric to measure the distance (similarity) of sample pairs. In this paper, we focus on the second one.

A large variety of pair-based losses have been proposed for deep metric learning. Contrastive loss \cite{hadsell2006dimensionality} explores the relationship between pairwise samples, i.e., minimizing the distance of positive pairs while maximizing the distance of negative pairs for being bigger than a margin. Triplet loss \cite{schroff2015facenet}, based on an anchor sample, aims to learn a distance metric by which the positive pair distances are smaller than the corresponding negative pair distances by a margin. Alternatively, lifted structured loss \cite{ohS2016deep,yang2018person} and N-pair loss \cite{sohn2016improved} are proposed to take the structure relationship of samples into consideration by assigning different weights for different pairs. Given a query sample, lifted structured loss \cite{ohS2016deep} intends to identify one positive sample from all corresponding negative samples, while N-pair loss \cite{sohn2016improved} intends to identify one positive sample from N-1 negative samples of N-1 classes.
Though driven by different motivations, these methods generally have a common principle of learning from informative pairs, by two steps (1) samples mining and (2) pairs weighting. Samples mining step selects the informative pairs out, and pairs weighting step assigns different weights for different pairs according to the pair distances since different pairs contribute differently.
Recently, ranked list loss \cite{wang2019ranked} and multi-similarity loss \cite{wang2019multi} are proposed to explicitly construct loss functions from the perspective of pairs weighting.
Based on contrastive loss \cite{hadsell2006dimensionality}, ranked list loss \cite{wang2019ranked} assigns weights for negative pairs based on the corresponding losses.
In \cite{wang2019multi}, under a general pairs weighing framework, multi-similarity loss function is proposed to fully consider three kinds of similarities for pairs weighting based on the pairs weighting strategy of lifted structured loss \cite{ohS2016deep}.

In this paper, motivated by ranked list loss \cite{wang2019ranked} and multi-similarity loss \cite{wang2019multi}, we unify those existing pair-based losses into a general pair-based weighting loss function, directly weighting the distances of informative pairs. We also focus on two aspects, samples mining for informative pairs selection and pairs weighting for taking advantage of different pairs. Those existing pair-based losses are reviewed inline with our general pair-based weighting loss formulation, attempting to deeply understand their motivations and analyse their adopting samples mining and pairs weighting methods. Based on the analysis, we explore some possible methods from the perspective of samples mining and pairs weighting. The potential combinations of these samples mining and pairs weighting methods are well considered. Finally, extensive experiments on three image retrieval benchmarks are conducted to evaluate our proposed general pair-based weighting loss functions.

To summarize, our main contributions are that.
\begin{itemize}
 \item We unify the pair-based loss functions into a general formulation, directly weighting the distances of informative pairs. It makes the design of pair-based loss function for deep metric learning to be two steps, samples mining and pairs weighting.
 \item Some possible samples mining and pairs weighting methods are detailedly explored, and the potential combinations of them are also considered.
\item Our general pair-based weighting loss function obtains new state-of-the-art performance on three standard image retrieval datasets.
\end{itemize}
%Multi-similarity loss \cite{wang2019multi}, Ranked list loss \cite{wang2019ranked}, HDML \cite{zheng2019hardness}, DAML \cite{duan2018deep}, Contrastive loss \cite{hadsell2006dimensionality}, Triplet loss \cite{schroff2015facenet}, Lifted structure loss \cite{ohS2016deep}, DDML \cite{hu2014discriminative}, N-pair loss \cite{sohn2016improved}, proxy-NCA \cite{movshovitz2017no}, Margin \cite{wu2017sampling}, HDC \cite{yuan2017hard}, Struct Clust \cite{oh2017deep}, HTL \cite{ge2018deep} ABIER \cite{opitz2018deep}, ABE \cite{kim2018attention}, Spectral clustering \cite{law2017deep}

\section{The general pair-based weighting loss}
In this section, we will first give the formulation of our general pair-based weighting loss function, and then review those existing pair-based losses inline with our general pair-based weighting loss formulation.

\subsection{The formulation of general pair-based weighting loss}
\label{ssec:gpblf}
Given paired data set $\{(x_i, x_j, y_{ij})\}$, where $y_{ij} \in \{0,1\}$ indicates whether a pair $(x_i, x_j)$ is from the same class ($y_{ij} = 1$) or not ($y_{ij} = 0$), the deep distance metric is defined as,
\begin{align}
    D(x_i, x_j) = d(z_i, z_j) = d\big(\theta; f(x_i), f(x_j)\big), \label{eq:dm}
\end{align}
where $z_i = f(x_i)$ is the learned embedding by the deep neural network with parameters $\theta$, and $d$ indicates the Euclidean distance $d(z_i, z_j) = \|z_i - z_j\|_2$.

Based on the deep neural network, deep metric learning (DML) methods are usually trained based on sample tuples composed of several samples with certain similarity relations, aiming to learn an embedding space, where the embedded vectors from the same class are encouraged to be closer, while those from different classes are pushed apart away from each other.
The deep neural network parameters $\theta$ are learned under a specifical loss objective. Based on the distance metric $D_{ij} = D(x_i, x_j)$ in Eqn. (\ref{eq:dm}), the minimizing optimization problem can be formulated as,
\begin{align}
\begin{tabular}{cl}
    $\theta^{*}$ & $= \underset{\theta}{\operatorname{argmin}} \,\,\,\, \mathcal{L} (\theta; \{D_{ij}\})$ \\
               & $= \underset{\theta}{\operatorname{argmin}} \,\,\,\, \mathcal{L} (\theta; \{D_{ij} | y_{ij} = 1\}) + \mathcal{L} (\theta; \{D_{ij} | y_{ij} = 0\})$,
\end{tabular}
\end{align}
where $\{D_{ij}\}$ is the whole paired distance set, $\{D_{ij} | y_{ij} = 1\}$ is the positive paired distance set, and $\{D_{ij} | y_{ij} = 0\}$ is the negative paired distance set.
%For simplification, in the following we will omit the $\theta$ in the loss functions.

The derivative with respect to neural network parameters $\theta$ can be calculated as,
\begin{align}
    \frac{\partial \mathcal{L} (\theta; \{D_{ij}\})}{\partial \theta} = \sum_{(i,j)} \frac{\partial \mathcal{L} (\theta; \{D_{ij}\})}{\partial D_{ij}} \,\,\,\, \frac{\partial D_{ij}}{\partial \theta}.
\end{align}

Note that when calculating the gradient of $\mathcal{L} (\theta; \{D_{ij}\})$ with respect to $\theta$, the term $\frac{\partial \mathcal{L} (\theta; \{D_{ij}\})}{\partial D_{ij}}$ is a constant number, which is only determined by the loss function, without the neural network parameters.
We term $w_{ij} = \frac{\partial \mathcal{L} (\theta; \{D_{ij}\})}{\partial D_{ij}}$. Therefore, the \textbf{general pair-based weighting loss function}\footnote{$D_{ij}$ may be with a bias term, such as $D_{ij} - m$.} can be uniformly reformulated as,
\begin{align}
    \mathcal{L} (\theta; \{D_{ij}\}) = \sum_{(i,j)} w_{ij} D_{ij}.
\end{align}

Since the core ideal of metric learning is to pull positive pairs as close as possible and push negative pairs apart from each other \footnote{Therefore, we assume that $\frac{\partial \mathcal{L} (\theta; \{D_{ij} | y_{ij} = 1\})}{\partial D_{ij}} \geq 0$, and $\frac{\partial \mathcal{L} (\theta; \{D_{ij} | y_{ij} = 0\})}{\partial D_{ij}} \leq 0$.}, we can simply reformulated the loss function in the following with two terms,
\begin{align}
    \mathcal{L} (\theta; \{D_{ij}\}) = \sum_{y_{ij} = 1} w_{ij} D_{ij} + \sum_{y_{ij} = 0} w_{ij} (-D_{ij}),  \label{eq:gpblf}
\end{align}
where $w_{ij} \geq 0$.

The aforementioned loss function is constructed on the pair sets $\{(x_i, x_j)\}$. However, some methods are based on triplet sets $\{(x_i, x_j, x_k) | y_{ij} = 1, y_{ik} = 0\}$ focusing on enlarging the margin between the positive and negative pair with an anchor. In this case, the minimizing optimization problem can be formulated as,
\begin{align}
    \theta^{*} & = \underset{\theta}{\operatorname{argmin}} \,\,\,\, \mathcal{L} (\theta; \{D_{ij}\}) \\ \nonumber
               & = \underset{\theta}{\operatorname{argmin}} \,\,\,\, \mathcal{L} (\theta; \{D_{ij} - D_{ik} | y_{ij} = 1, y_{ik} = 0\}).
\end{align}

The derivative with respect to neural network parameters $\theta$ can be calculated as,
\begin{align}
    \frac{\partial \mathcal{L} (\theta; \{D_{ij}\})}{\partial \theta} & \label{eq:lijkp} \\  \nonumber
     = \sum \limits_{\substack{(i,j,k), \\ y_{ij} = 1,y_{ik} = 0}} & \frac{\partial \mathcal{L} (\theta; \{D_{ij} - D_{ik}\})}{\partial (D_{ij} - D_{ik})} \,\,\, \frac{\partial (D_{ij} - D_{ik})}{\partial \theta} ,
\end{align}
where
\begin{align}
    \frac{\partial (D_{ij} - D_{ik})}{\partial \theta} \quad \,\,&  \label{eq:dijkp}  \\   \nonumber
        =  \frac{\partial (D_{ij} - D_{ik})} {\partial D_{ij}} \,\,& \,\,\, \frac{\partial D_{ij}}{\partial \theta}
        +  \frac{\partial (D_{ij} - D_{ik})}{\partial D_{ik}}  \,\,\, \frac{\partial D_{ik}}{\partial \theta} \\
        =  \frac{\partial D_{ij}}{\partial \theta} \quad \,\,\,\, + \quad \,\, & \frac{\partial (-D_{ik})}{\partial \theta}. \nonumber
\end{align}

Similarly, we term $w_{ijk} = \frac{\partial \mathcal{L} (\theta; \{D_{ij} - D_{ik}\})}{\partial (D_{ij} - D_{ik})}$ \footnote{Similar to $w_{ij}$, we also can assume $w_{ijk} = \frac{\partial \mathcal{L} (\theta; \{D_{ij} - D_{ik}\})}{\partial (D_{ij} - D_{ik})} \geq 0$ for $\{(i,j,k) |  y_{ij} = 1,y_{ik} = 0\}$.}. When combining Eqn. (\ref{eq:dijkp}) into Eqn. (\ref{eq:lijkp}), we can obtain the following,
\begin{align}
     \frac{\partial \mathcal{L} (\theta; \{D_{ij}\})}{\partial \theta} &  \\ \nonumber
     = \sum \limits_{\substack{(i,j,k), \\ y_{ij} = 1,y_{ik} = 0}} & w_{ijk} \,\,\, \frac{\partial (D_{ij} - D_{ik})}{\partial \theta} \\ \nonumber
     = \sum \limits_{\substack{(i,j,k), \\ y_{ij} = 1,y_{ik} = 0}} & w_{ijk} \,\,\, (\frac{\partial D_{ij}}{\partial \theta} +  \frac{\partial (-D_{ik})}{\partial \theta}).
\end{align}

Consequently, the triplets based loss function can be reformulated as,
\begin{align}
    \mathcal{L} (\theta; \{D_{ij}\}) = \sum \limits_{\substack{(i,j,k), \\ y_{ij} = 1,y_{ik} = 0}} w_{ijk} (D_{ij} - D_{ik}),  \label{eq:gtblf}
\end{align}
where $w_{ijk} \geq 0$.

Compare our triplet-based loss function (Eqn. (\ref{eq:gtblf})) to our pair-based loss function (Eqn. (\ref{eq:gpblf})), we can find that the triplet-based loss function is essentially the same to the pair-based loss function, assigning different weights for each pair (positive or negative). The only difference is that triplet-based loss function assigns equal weight for both positive and negative pairs in a triplet.

From Eqn. (\ref{eq:gpblf}) and Eqn. (\ref{eq:gtblf}), we can know that there are two main points the loss functions should focus on.
\begin{enumerate}
\item \textbf{Samples mining to construct the informative positive and negative pair sets, or reform into the triplet sets.}
\item \textbf{The design of pair-based loss function is essentially computing the weight for each pair.}
\end{enumerate}

\subsection{Review the existing pair-based loss functions}
In this section, we will briefly review some existing pair-based loss functions and unify them into our general pair-based weighting loss function, including contrastive loss \cite{hadsell2006dimensionality}, triplet loss \cite{schroff2015facenet}, N-pair loss \cite{sohn2016improved}, lifted structured loss \cite{ohS2016deep}, and multi-similarity loss \cite{wang2019multi}. For simplification, in the following we will discard the parameters $\theta$ when describing the loss functions.
\subsubsection{Contrastive loss}
Based on paired data, contrastive loss \cite{hadsell2006dimensionality} is designed to minimize the distance \footnote{Note that, for both contrastive loss \cite{hadsell2006dimensionality} and triplet loss \cite{schroff2015facenet}, in the original paper, they focus on the square of pair distance. The difference of distance and square distance in loss function, we will discuss in Section \ref{sssec:sdis} and \ref{sssec:pairw}.} between a positive pair while penalize the distance between a negative pair being smaller than a predefined margin $m$,
\begin{align}
    \mathcal{L} (\{D_{ij}\}) = \sum_{y_{ij} = 1} D_{ij} + \sum_{ y_{ij} = 0}[m - D_{ij}]_{+},
\end{align}
where $[\cdot]_{+}$ operation indicates the hinge function $max(0,\cdot)$.
Concretely, contrastive loss (1) only mines negative pairs whose distance is smaller than $m$, and (2) assigns equal weight ($w_{ij} = 1$) for all the selected pairs.

\subsubsection{Triplet loss}
Based on triplet data $\{(x_i,x_j,x_k), | y_{ij} = 1, y_{ik} = 0\}$, triplet loss \cite{schroff2015facenet} is designed to make the distance of positive pair be smaller than the distance of negative pair by a margin $m$ in a triplet,
\begin{align}
    \mathcal{L} (\{D_{ij}\}) = \sum \limits_{\substack{(i,j,k), \\ y_{ij} = 1, y_{ik} = 0}} [D_{ij} - D_{ik} + m]_{+}.
\end{align}

Concretely, triplet loss (1) mines suitable samples to form informative triplets, and (2) also assigns equal weight ($w_{ijk} = 1$) for all the selected pairs.

\subsubsection{N-pair loss }
Triplet loss pulls one positive sample while pushing a negative one simultaneously. To improve the triplet loss by interacting with more negative classes and samples, N-pair loss \footnote{N-pair loss \cite{sohn2016improved} and multi-similarity loss \cite{wang2019multi} are originally described in the similarity formulation, which is essentially the same to the Euclidean distance formulation, just as the L2 normalized Euclidean distance vs. cosine similarity.} \cite{sohn2016improved} is designed to identify one positive sample from $N - 1$ negative samples of $N - 1$ classes (one negative sample per class),
\begin{align}
    \mathcal{L} (\{D_{ik}\}) = \sum_{y_{ii} = 1} \log \Big(1 + \sum_{y_{ik} = 0}\exp(D_{ii} - D_{ik})\Big),
\end{align}
where $\{D_{ik} \big| i,k = 1,\cdots,N;  y_{ii} = 1; \text{and}\,\, y_{ik} = 0, \text{if}\,\, k \neq i\}$, and $D_{ii}$ is the distance of a positive pair $x_i$ and $x_{i}^{+}$.

Concretely, N-pair loss (1) mines the informative samples by randomly sampling one positive pair for each class to form N pairs, then utilizing the multi-class relationships among the N pairs. (2) Then it assigns equal weight ($w_{iik} = \frac{\exp(D_{ii} - D_{ik})}{1+\sum_{k \neq i}\exp(D_{ii} - D_{ik})}$) for both the positive and negative pair in a triplet.

\subsubsection{Lifted structured loss}
Instead of only utilizing one negative sample of each class,  lifted structured loss \cite{ohS2016deep} is designed to incorporating all the negative samples in the training mini-batch. For each positive pair $(x_i, x_j)$, lifted structured loss aims to pull $(x_i$ and $x_j)$ as close as possible and push all the negative samples corresponding to $(x_i$ and $x_j)$ farther than a margin $m$, which can be mathematically formulated as,
\begin{align}
    \mathcal{L} (\{D_{ij}\}) =& \sum_{y_{ij} = 1}\Big[D_{ij} + \log\Big(\sum_{y_{ik}=0}\exp(m - D_{ik})\Big) \\
    & + \log \Big(\sum_{y_{jl}=0}\exp(m - D_{jl})\Big)\Big]_{+}. \nonumber
\end{align}
Given a query sample, lifted structured loss intends to identify one positive sample from all corresponding negative samples in the mini-batch to full explore the structure relationship. Concretely, lifted structured loss (1) utilizes all the negative samples and mines informative samples which make the hinge function return a non-zero value. Then (2) for the positive pair, lifted structured loss assigns weight $w_{ij} = 1$, while assigns different weights for each negative pair $w_{ik} = \frac{\exp(m - D_{ik})}{\sum_{y_{ik}=0}\exp(m - D_{ik})}$.

\subsubsection{Multi-similarity loss}
Based on lifted structured loss, multi-similarity loss \cite{wang2019multi} is designed to full utilize both positive and negative pairs by a more generalized weighting strategy. Firstly, by a predefined threshold $\epsilon$, it mines informative pairs by the triplet criterion: those positive pairs are sampled whose distance are bigger than the smallest distance of negative pairs minimizing $\epsilon$ (Eqn. (\ref{eq:tripmp})), while those negative pairs are sampled whose distances are smaller than the biggest distance of positive pairs adding $\epsilon$ (Eqn. (\ref{eq:tripmn})). Then the loss is calculated as,
\begin{align}
    \mathcal{L} (\{D_{ij}\}) &= \frac{1}{\alpha} \log \Big[1+ \sum_{y_{ij} = 1}\exp \big(\alpha(D_{ij} - m)\big)\Big] \\
     &+ \frac{1}{\beta} \log \Big[1+ \sum_{y_{ik} = 0}\exp \big(\beta(m - D_{ik})\big)\Big], \nonumber
\end{align}
where $\alpha$, $\beta$ and $m$ are predefined hyper-parameters to well control the weights for different pairs.
Concretely, multi-similarity loss (1) mines informative pairs by the triplet criterion for both positive and negative pairs, then (2) respectively assigns different weights, $w_{ij} = \frac{\exp \big(\alpha(D_{ij} - m)\big)}{1+\sum_{y_{ij} = 1}\exp \big(\alpha(D_{ij} - m)\big)}$ for positive pairs, while $w_{ik} = \frac{\exp \big(\beta(m - D_{ik})\big)}{1+\sum_{y_{ik} = 0}\exp \big(\beta(m - D_{ik})\big)}$ for negative pairs.

Comparing the weights of multi-similarity (MS) loss to the weights of lifted structured (LS) loss, we can find that there are three differences. (1) ML loss computes weights for both of the positive and negative pairs, while LS loss only focuses on negative pairs. (2) The weights of ML loss adds 1 in the denominator. In my opinion, this operation maybe meaningless because any sample pair $D_{ij} \geq m$ will make the denominator of $w_{ij}$ increasing more than 1, also $D_{ik} \leq m$ to $w_{ik}$. The parameter $\beta$ also will affect the values. (3) ML loss adds the temperature parameters $\alpha$ and $\beta$ to well adjust the weights.

Through the above analysis, we can unify these above pair-based loss functions into our general loss functions (pair-based weighting loss function (Eqn. (\ref{eq:gpblf})) or triplet-based weighting loss function (Eqn. (\ref{eq:gtblf}))). Among them, contrastive loss \cite{hadsell2006dimensionality}, lifted structured loss \cite{ohS2016deep}, and multi-similarity loss \cite{wang2019multi} are in line with pair-based weighting loss function (Eqn. (\ref{eq:gpblf})), while triplet loss \cite{schroff2015facenet} and N-pair loss \cite{sohn2016improved} are inline with triplet-based weighting loss function (Eqn. (\ref{eq:gtblf})).
\begin{figure}
\centering
\includegraphics[width=90mm]{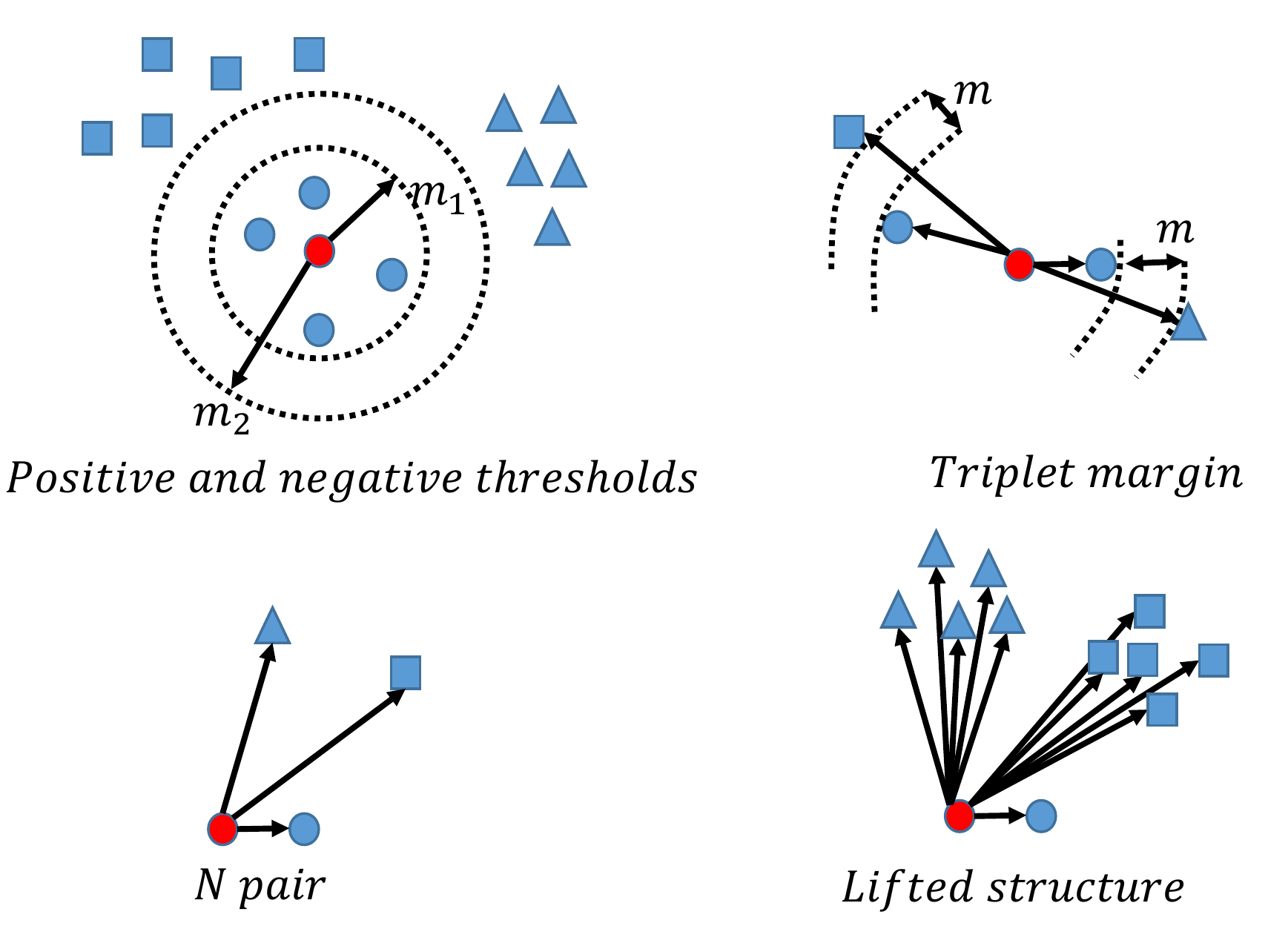}
\caption{The illustration of different samples mining methods. We take 3 classes for an example, where different shapes indicate different classes, and the red circle is an anchor.}
\label{fig:sample_illus}
\end{figure}

\section{Methodologies}
In this section, based on our general weighting loss functions (Eqn. (\ref{eq:gpblf}) and Eqn. (\ref{eq:gtblf})), we will explore some possible methods from the perspective of samples mining and pairs weighting.
\subsection{Samples mining}
There are two places referring to the samples mining. One is the sampler strategy to construct the mini-batch during training. The other is the informative samples mining in the loss function.

\noindent\textbf{Mini-batch construction.} There are two ways to construct the a mini-batch. (1) Random sampler. In this case, all samples and classes are randomly sampled. (2) $PK$ sampler, randomly sampling $P$ classes and $K$ images per class. In this case, in a mini-batch, for every query image there are $K-1$ positive images and $(P-1)K$ negative images. It can simulate the true structure of real data.

Here, we concentrate on the second aspect, how to mine the informative pairs for loss functions in a mini-batch. The details are illustrated in Fig. \ref{fig:sample_illus}.

\subsubsection{Positive and negative thresholds}
Since the core ideal of metric learning is to pull positive pairs as close as possible and push negative pairs apart from each other, the intuitive insight is to set thresholds. The distances of negative pairs should be larger than a predefined threshold, while the distances of positive pairs should be smaller than a predefined threshold. Those pairs which are not meet the criterion should be pay more attention to. Mathematically, it can formulated as,
\begin{align}
    \{D_{ij}^{+} \quad |\quad  D_{ij} \geq m_1, y_{ij} = 1 \}, \label{eq:smpp}\\
    \{D_{ij}^{-} \quad |\quad  D_{ij} \leq m_2, y_{ij} = 0 \}, \label{eq:smpn}
\end{align}
where $0 \leq m_1 \leq m_2$. In this case, the general pair-based weighting loss function (Eqn. (\ref{eq:gpblf})) could be reformulated as,
\begin{align}
    \mathcal{L} (\{D_{ij}\}) = \sum_{y_{ij} = 1} w_{ij} [D_{ij} - m_1]_{+} + \sum_{y_{ij} = 0} w_{ij} [m_2 - D_{ij}]_{+}. \label{eq:gpblf_}
\end{align}
Contrastive loss \cite{hadsell2006dimensionality}, discriminative deep metric learning \cite{hu2014discriminative}, margin  based loss \cite{wu2017sampling} and ranked list loss \cite{wang2019ranked} are all in this samples mining manner. The only difference is with different threshold settings for $m_1$ and $m_2$.

\subsubsection{Triplet margin}
Positive and negative thresholds are only focusing on the self pair distances, ignoring the relative pair distances. Therefore, triplet margin is designed to select those samples in a triplet, where the distance of positive pair adding a margin $m$ is bigger than that of negative pair. It can be mathematically formulated as,
\begin{align}
    \{(D_{ij},D_{ik}) | D_{ij} + m \geq D_{ik}, y_{ij} = 1,  y_{ik} = 0\},
\end{align}
where $m \geq 0$.

For calculation simplicity, Hermans et al. \cite{hermans2017defense} proposed an organizational modification to mine the hardest triplets.
\begin{align}
    \{(D_{ij},D_{ik}) | D_{ij} = \max_{y_{ij} = 1} D_{ij}, D_{ik} = \max_{y_{ik} = 0} D_{ik}\}.
\end{align}

In this case, the general triplet-based weighting loss function (Eqn. (\ref{eq:gtblf})) could be reformulated as,
\begin{align}
    \mathcal{L} (\{D_{ij}\}) = \sum \limits_{\substack{(i,j,k), \\ y_{ij} = 1, y_{ik} = 0}} w_{ijk} [D_{ij} - D_{ik} + m]_{+}. \label{eq:gtblf_}
\end{align}

Moreover, based on the triplet margin method, we also can mine the positive and negative pairs in a mini-batch as done by multi-similarity loss \cite{wang2019multi}.
\begin{align}
    \{D_{ij}^{+} \quad | \quad D_{ij} \geq (\min_{y_{ik = 0}} D_{ik}) - \epsilon , y_{ij} = 1 \}, \label{eq:tripmp}\\
    \{D_{ik}^{-} \quad | \quad D_{ik} \leq (\max_{y_{ij = 1}} D_{ij}) + \epsilon , y_{ik} = 0 \}. \label{eq:tripmn}
\end{align}
This samples mining method is similar to the aforementioned positive and negative thresholds method. Positive and negative thresholds method sets two fixed thresholds $m_1$ and $m_2$, where $0 \leq m_1 \leq m_2$, while the this method sets one margin $\epsilon$ for dynamically controlling the hardest informative samples.

\subsubsection{Structure relationship mining}
To effectively utilize the information among the training samples in a mini-batch, the structure relationships are explored.
Multi-class N-pair loss \cite{sohn2016improved} incorporates more negative classes. It samples a positive pair from each class (N pairs), aiming to identify one positive sample
from N-1 negative samples of N-1 classes (one negative sample per class).
Moreover, for every positive pair ($x_i$ and $x_j$), lifted structured loss \cite{ohS2016deep} takes all the negative pairs, corresponding to $x_i$ and $x_j$, into account in a mini-batch, aiming to identify one positive sample from all corresponding negative samples.
In fact, the $PK$ sampler for mini-batch construction is already including the structure relationship. These two structure relationship mining methods can be simply simulated by setting $K=2$.

%The structure relationship mining methods are always working with the corresponding weighting methods proposed in
\begin{figure}[t]
\centering
\begin{tabular}{cc}
\includegraphics[width=4.2cm]{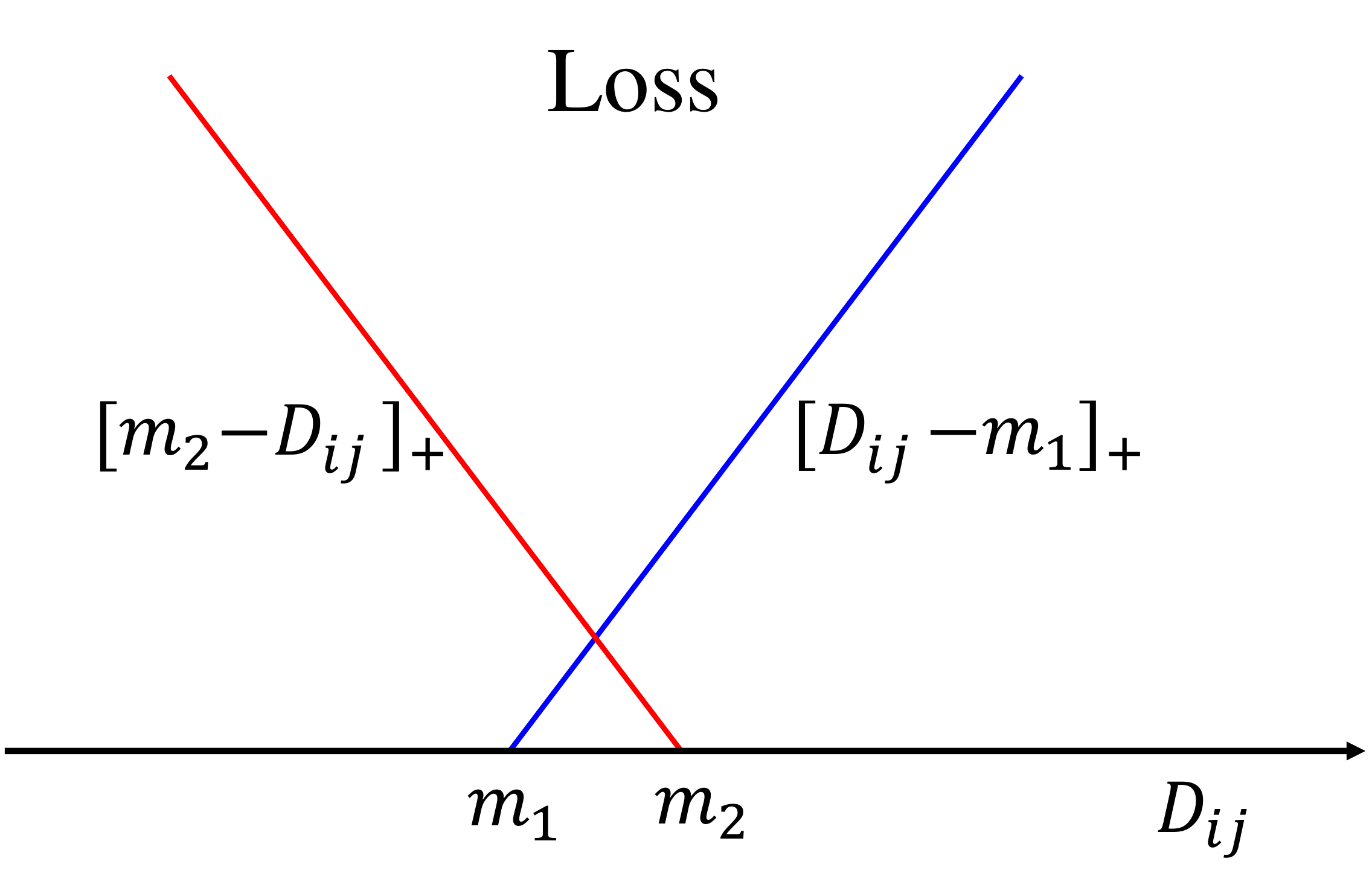} & \includegraphics[width=4.2cm]{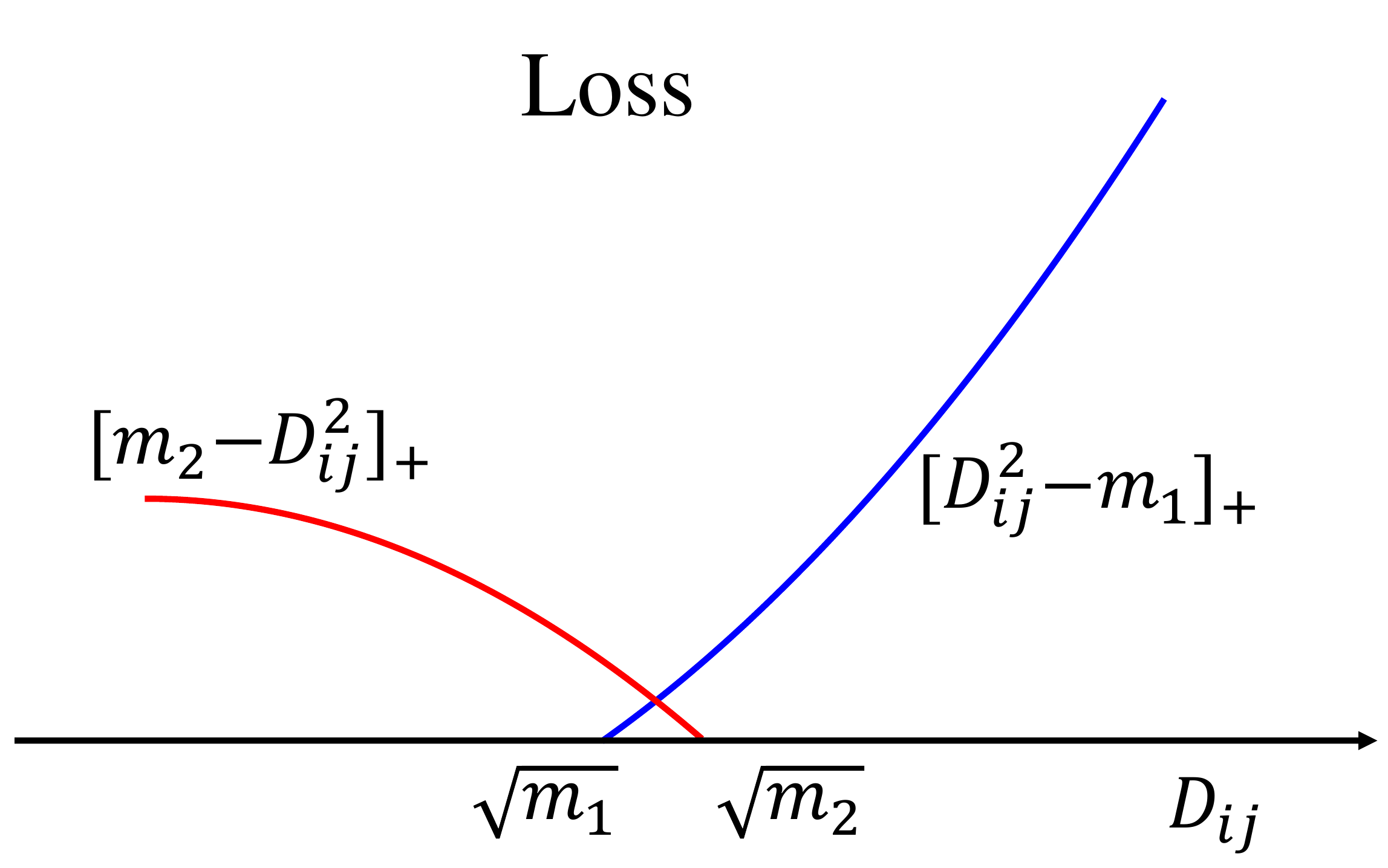}\\
(a)&(b)\\
\includegraphics[width=4.2cm]{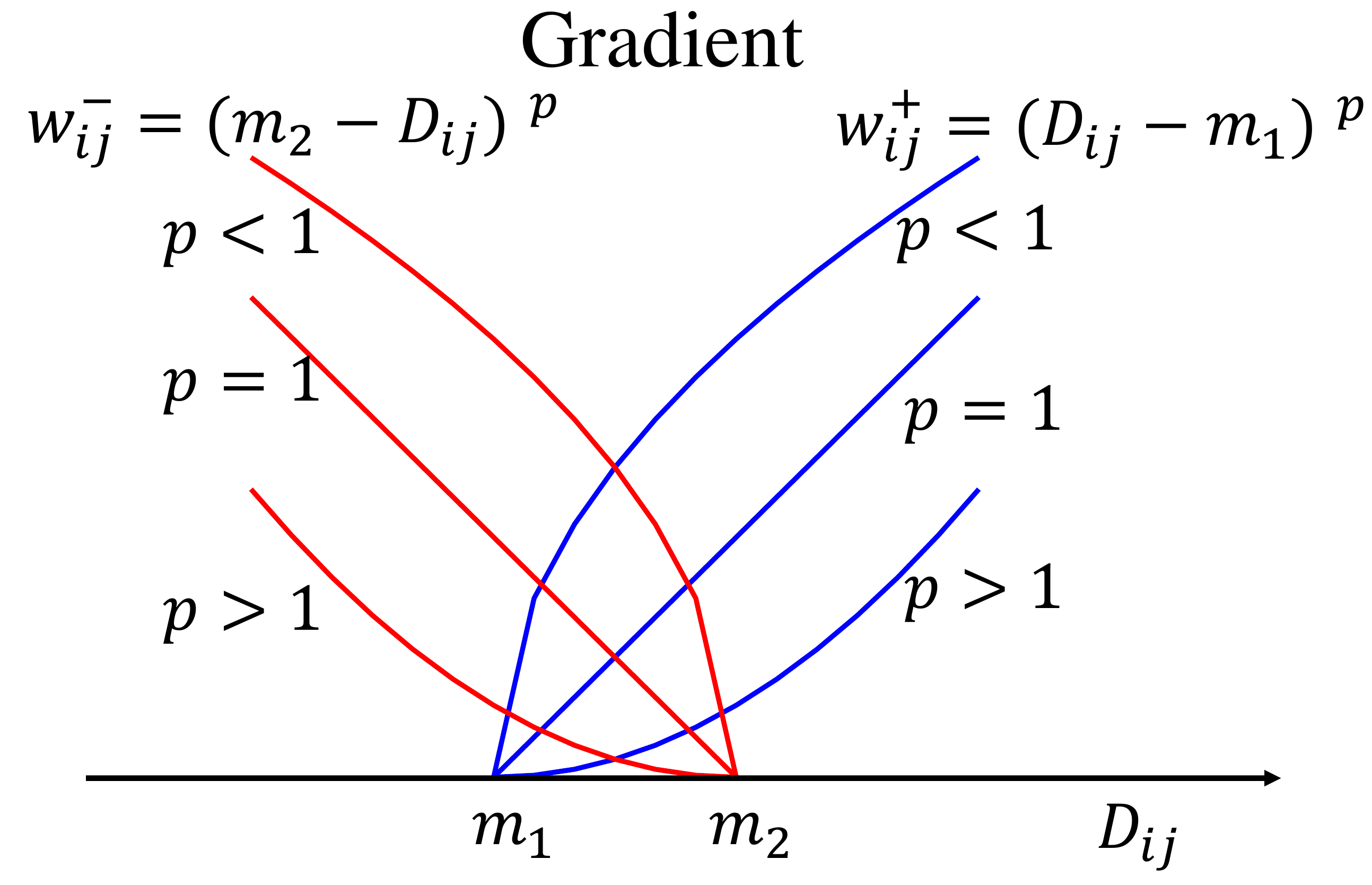} & \includegraphics[width=4.2cm]{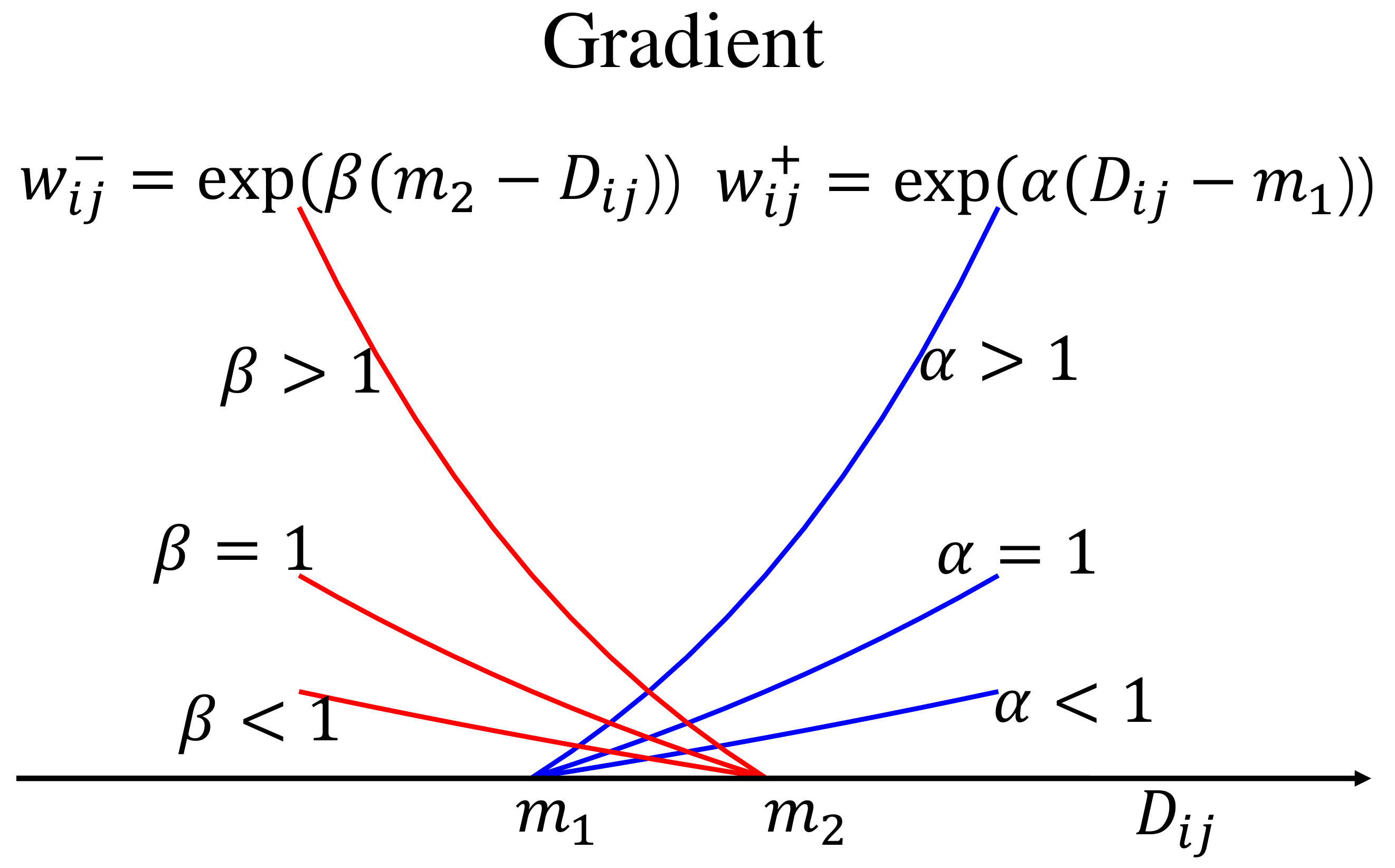}\\
(c)&(d)\\
\end{tabular}
\caption{The illustration of different weight strategies based on the pair-based loss functions, blue curves for positive pairs and red curves for negative pairs. (a) The contrastive loss based on distances. (b) The  contrastive loss based on square distances. (c) The $p_{th}$ power weights. (d) The exponential weights.}
\label{fig:weight_comp}
\end{figure}

\subsection{Pairs weighting}
In view of the general pair-based weighting loss function (Eqn. (\ref{eq:gpblf_})) and triplet-based weighting loss function (Eqn. (\ref{eq:gtblf_})), after the samples mining, the most important step is the weighting for selected pairs. Different pairs have different influences on the loss function through assigning different weights for gradient computation.
Here, we explore some possible weighting strategies. Note that, the weight $w_{ij}$ in our loss functions are just the calculated number without gradient\footnote{We compute the weight $w_{ij}$ with the operation of $D_{ij}.detach()$ in Pytorch framework.}.
We term $w_{ij}^{+}$ as the weight for positive pair, and $w_{ij}^{-}$ as the weight for negative pair.

\subsubsection{Constant weight}
The most simple case is treat all the selected pairs equally with a constant weight.
For pair-based loss,
\begin{align}
    w_{ij}^{+} = w_{ij}^{-} = 1.
\end{align}
For triplet-based loss,
\begin{align}
    w_{ijk} = 1.
\end{align}

Contrastive loss \cite{hadsell2006dimensionality}, triplet loss \cite{schroff2015facenet} and margin based loss \cite{wu2017sampling} are in this case.

\subsubsection{The $p_{th}$ power weight}
Based on the pair distances, we can assign different weights for each pair. The core principle is that, (1) for positive pair, the bigger the distance is the bigger the weight should be. (2) For negative pair, the smaller the distance is the bigger the weight should be. Big weight denotes big absolute gradient, corresponding to the hard sample.

The $p_{th}$ ($p\geq0$) power could be adopted to calculate weight. For pair-based loss,
\begin{align}
    w_{ij}^{+} = (D_{ij} - m_1)^{p},  \label{eq:powerp}
\end{align}
\begin{align}
    w_{ij}^{-} = (m_2 - D_{ij})^{q}, \label{eq:powern}
\end{align}
where $p$ and $q$ are predefined hyper-parameters to well control the weights for different pairs.

For triplet-based loss,
\begin{align}
    w_{ijk} = (D_{ij} - D_{ik} + m)^{p}.  \label{eq:powert}
\end{align}
When $p = q = 0$, the  $p_{th}$ power weight is just the aforementioned constant weight $w_{ij}=1$.

\subsubsection{Exponential weight}
When computing the weight, the exponential operation is frequently utilized, such as multi-similarity loss \cite{wang2019multi}, ranked list loss \cite{wang2019ranked}, hardness-aware deep metric learning \cite{zheng2019hardness}, deep adversarial metric learning \cite{duan2018deep}, lifted structure loss \cite{ohS2016deep}, N-pair loss \cite{sohn2016improved} and discriminative deep metric learning \cite{hu2014discriminative}.

For pair-based loss,
\begin{align}
    w_{ij}^{+} = \exp(\alpha(D_{ij} - m_1)),  \label{eq:exponentialp}
\end{align}
\begin{align}
    w_{ij}^{-} = \exp(\beta(m_2 - D_{ij})).   \label{eq:exponentialn}
\end{align}
For triplet-based loss,
\begin{align}
    w_{ijk} = \exp(\alpha(D_{ij} - D_{ik} + m)),  \label{eq:exponentialt}
\end{align}
where $\alpha$ and $\beta$ are predefined hyper-parameters to well control the weights for different pairs. When $\alpha = \beta = 0$, the exponential weight is just the aforementioned constant weight $w_{ij}=1$.

\begin{figure}
\centering
\includegraphics[width=90mm]{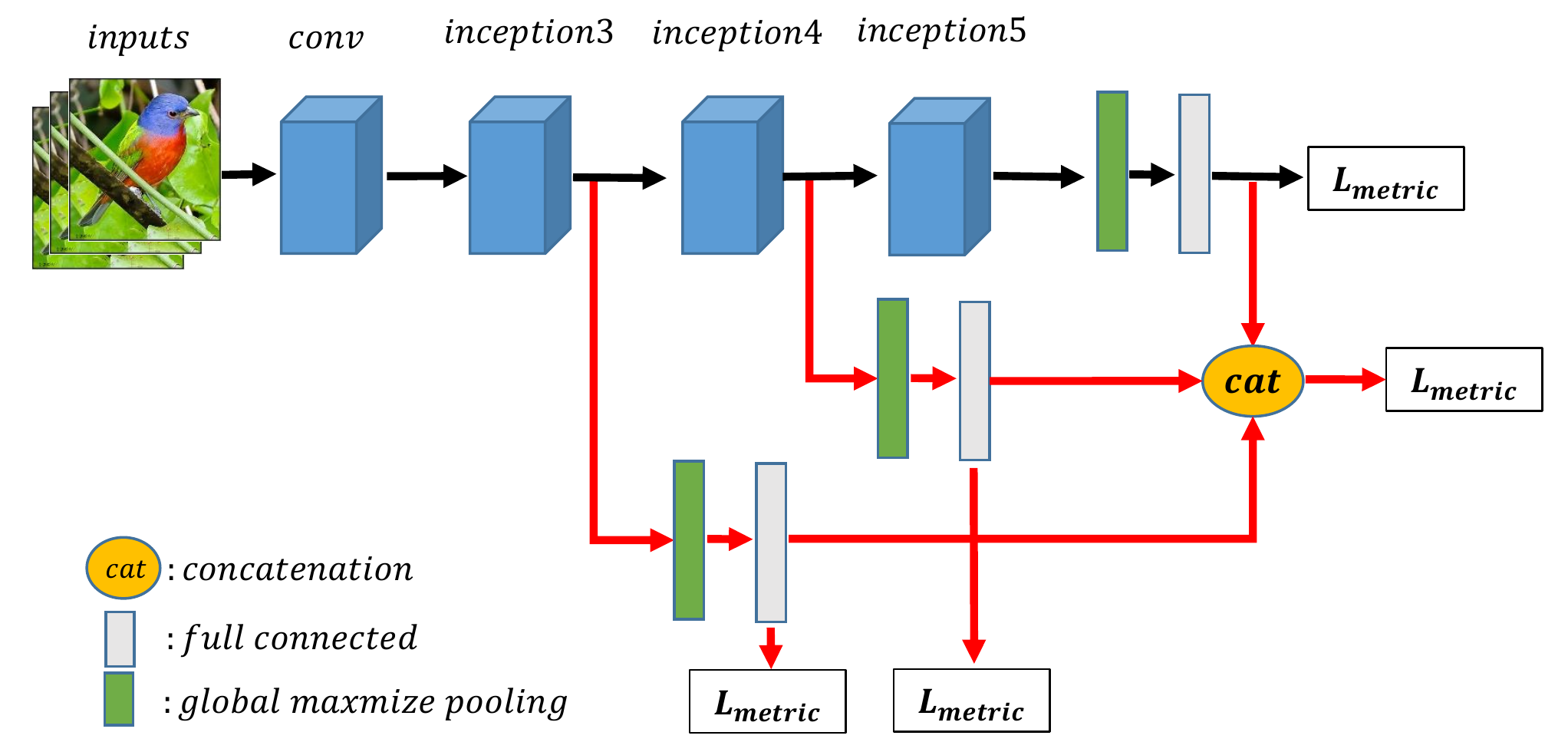}
\caption{The framework for our deep metric learning. We take the bn-inception model as the backbone. The inputs images are fed into the network to extract features, guided by the metric loss. The black lines are the flows of baseline architecture, while adding those red lines are the multi-level features fusion.}
\label{fig:framework}
\end{figure}

\subsubsection{Weight normalization}
\label{sssec:wnorm}
The above $p_{th}$ power weight and exponential weight only focus on the self pair distance, ignoring the relative pair distance which, however, could reflect the structure information of training samples in a mini-batch. Therefore, to well utilized the structure relationship of pairs in a mini-batch, we respectively \textbf{normalize} those pair weights.

For pair-based loss,
\begin{align}
    w_{ij}^{+} = \frac{w_{ij}^{+}}{\sum_{y_{ij=1}}w_{ij}^{+}},
\end{align}
\begin{align}
    w_{ij}^{-} = \frac{w_{ij}^{-}}{\sum_{y_{ij=0}}w_{ij}^{-}}.
\end{align}
For triplet-based loss,
\begin{align}
    w_{ijk} = \frac{w_{ijk}}{\sum_{y_{ij=1,ik=0}}w_{ijk}}.
\end{align}

\subsubsection{The square distance}
\label{sssec:sdis}
As aforementioned, the contrastive loss \cite{hadsell2006dimensionality} and triplet loss \cite{schroff2015facenet} are implemented in square distance formulation.
The original contrastive loss \cite{hadsell2006dimensionality},
\begin{align}
    \mathcal{L} (\{D_{ij}\}) = \sum_{y_{ij} = 1} [D_{ij} - m_1]_{+}^{2} + \sum_{y_{ij} = 0} [m_2 - D_{ij}]_{+}^{2},
\end{align}
which is essentially the same to our $p_{th}$ power weight when $p = 1$.

However, we can use the square distance in another manner,
\begin{align}
    \mathcal{L} (\{D_{ij}\}) = \sum_{y_{ij} = 1} [D_{ij}^{2} - m_1]_{+} + \sum_{y_{ij} = 0} [m_2 - D_{ij}^{2}]_{+}.  \label{eq:squarep}
\end{align}

The original triplet loss \cite{schroff2015facenet},
\begin{align}
    \mathcal{L} (\{D_{ij}\}) = \sum \limits_{\substack{(i,j,k), \\ y_{ij} = 1, y_{ik} = 0}} [D_{ij}^{2} - D_{ik}^{2} + m]_{+}.  \label{eq:squaret}
\end{align}

These two loss functions based on square distance will lead to different kinds of gradients compared to our $p_{th}$ power weight method.

\begin{algorithm}[t]
    \caption{The general pair-based weighting loss for deep metric learning.}
    \label{alg:gpl}
    \KwIn{All the training images; the pre-trained network parameters $\theta$; the parameters for loss functions: $m_1$, $m_2$, $\epsilon$, $p$, $q$, $\alpha$, $\beta$;}

    \KwOut{The updated network parameters $\theta$;}
    \Begin{Iterate step1 to step6:\\
        Step 1: Construct the mini-batch $X = \{x_i\}$ by $PK$ sampler;\\
        Step 2: Feed $X$ into the network to obtain embeddings $Z = \{f(\theta;x_i)\}$;\\
        Step 3: Compute the pair distance $D = \{\|z_i - z_j\|_2 \}$ for $Z$;\\

        Step 4: Iteration\\
                \For {each $z_i \in Z$}
                {Sample the positive pairs and negative pairs based on $D$ according to Eqns. (\ref{eq:smpp}, \ref{eq:smpn});\\
                 Compute the weights for every pair according to Eqns. (\ref{eq:powerp}, \ref{eq:powern}, \ref{eq:powert}) or Eqns. (\ref{eq:exponentialp}, \ref{eq:exponentialn}, \ref{eq:exponentialt}); \\
                 Compute the loss $\mathcal{L}_i$ for $z_i$ according to Eqn. (\ref{eq:gpblf_});\\
                }
        Step 5: Calculate the final loss for mini-batch $Z$, $\mathcal{L} = \frac{1}{PK} \sum_{i}^{PK} \mathcal{L}_i$; \\
        Step 6: Perform backpropagation to update the network parameters $\theta$. \\
        \Return{The updated network parameters $\theta$.}
    }
 \end{algorithm}

\subsubsection{The comparison of different weights strategies}
In our pair-based loss function, those negative pairs with small distances will lead large losses, while those positive pairs with large distances will lead large losses. Different loss functions are with different magnitudes for each pair through the weights (also the gradients).

We compare the difference of distance and square distance in contrastive loss, as shown in Fig. \ref{fig:weight_comp} (a) and (b). The distance loss treats every  pair equally, with gradients $|w_{ij}| = 1$ for every $D_{ij}$. However, the square distance loss is with gradients $|w_{ij}| = 2D_{ij}$, which will give zero gradients to those negative pairs with zero distance.

The $p_{th}$ power and exponential weight strategies are shown in Fig. \ref{fig:weight_comp} (c) and (d). We can see that they truly satisfy the core principle for weighting. (1) For positive pair, the bigger the distance is the bigger the weight should be. (2) For negative pair, the smaller the distance is the bigger the weight should be.
The magnitudes of weights can be adjusted by the hyper-parameters to well utilize different pairs.

\section{Experiments}

In this section, we conduct experiments to evaluate the effectiveness of our general pair-based weighting loss for deep metric learning on three standard datasets, CUB-200-2011 \cite{wah2011caltech}, Cars-196 \cite{krause20133d} and  Stanford online products (SOP) \cite{ohS2016deep}.
\subsection{Experimental settings}
\noindent\textbf{Datasets.}
Following the setup of \cite{ohS2016deep,wang2019multi,wang2019ranked,zheng2019hardness}, we perform the image retrieval task under zero-shot settings, where the training set and test set contain image classes with no intersection.

(1) CUB-200-2011 \cite{wah2011caltech} consists of 11,788 images of 200 bird species.  5,864 images of the first 100 classes are used for training and 5,924 images of the other 100 classes for testing.

(2) Cars-196 \cite{krause20133d} has 16,185 images of 196 car model categories. We use the first 98 classes (8,054 images) for training and the remaining 98 classes (8,131 images) for testing.

(3) Stanford online products (SOP) \cite{ohS2016deep} is composed of 120,053 images of 22,634 online products sold on eBay.com. 59,551 images of 11,318 categories and 60,502 images of 11,316 categories are used for training and testing respectively.

\noindent\textbf{Architecture.} For fair comparison with \cite{wang2019multi,wang2019ranked}, we adopt the GoogleNet V2 (BN-inception) \cite{ioffe2015batch} as our backbone network. The details are shown in Fig. \ref{fig:framework}.
(1) We add three fully connected layers after $inception3$, $inception4$ and $inception5$ modules, respectively.
(2) The black lines are the flows of baseline architecture.
(3) Adding those red lines are the multi-level features fusion, by $cat$ operation.

\begin{table}
\caption{The results of baseline on the CUB-200-2011.}
\label{tab:baseline}
  \centering
  \begin{tabular}{l|c|c|c|c|c|c}
    \toprule[2pt]
    Racall@K (\%)  & 1 & 2 & 4 & 8 & 16 & 32 \\ \toprule[1pt]
     no freeze BN & 61.2 & 73.3  & 82.7 & 89.7 & 94.2 & 96.9 \\
     no freeze backbone & 65.2 & 76.3 & 84.0 & 90.4 & 94.5 & 96.9 \\
     baseline(pair-based) & 65.5 & 76.6 & 84.7 & 90.8 & 94.3 & 97.3 \\ \toprule[1pt]
     baseline(triplet-based) & 64.3 & 75.8 & 84.5 & 90.4 & 95.0 & 97.3 \\ \toprule[2pt]
  \end{tabular}
\end{table}

\noindent\textbf{Metrics.}  We report the image retrieval performance in terms of Recall$@K$, which is determined by the existence of at least one correct retrieved sample in the $K$ nearest neighbors.  All the features are $L2$ normalised before computing their distance during training and testing.

\noindent\textbf{Implementation details.} We implement all the methods in Pytorch framework. The procedures of our general pair-based loss function for deep metric learning is shown in Algorithm \ref{alg:gpl}. In the experiments, we adopt the $PK$ sampler to construct mini-batch, setting $K=5$, and $P=16$ for CUB-200-2011 \cite{wah2011caltech} and Cars-196 \cite{krause20133d} datasets, $P=64$ for Stanford online products \cite{ohS2016deep} dataset. All the input images were cropped to $224 \times 224$. For data augmentation, we used random crop with random horizontal mirroring for training, and single center crop for testing. Adam optimizer was used for all experiments with a learning rate 1e-5. The pre-trained model on ImageNet \cite{russakovsky2015imagenet} is used for initialisation. According to \cite{ohS2016deep}, the new fully connected layers are randomly initialised and optimised with 10 times larger learning rate than the others for faster convergence.
Moreover, we freeze all the batch normalization layers during training and freezing the BN-inception backbone network for the first 5 epochs.
For CUB-200-2011, training lasts for 20 epochs. For Cars-196 and stanford online products (SOP), training lasts for 60 epochs.

\subsection{Ablation study}
To demonstrate the effectiveness of our general pair-based weighting loss function (Eqns. (\ref{eq:gpblf_}) and (\ref{eq:gtblf_})) for deep metric learning, we conduct ablation experiments on CUB-200-2011. The output dimensions of fully connected layers are set  $d=512$.
\subsubsection{A strong baseline}
We adopt the baseline architecture (without multi-level features fusion), and ignore the weight in our general pair-based loss function (Eqn (\ref{eq:gpblf_})), setting $w_{ij} = 1$ for all pairs. We empirically set $m_1 = 0$ and $m_2 = 0.8$. During training, we add two tricks to boost the performance. (1) Freezing all the batch-normalization layers. (2) Freezing the BN-inception backbone network for 5 epochs, namely only training the newly added fully connected layers for the first 5 epochs.
The results are listed in Table \ref{tab:baseline}. We can find that (1) freezing the batch-normalization (BN) layers truly can greatly improve the performance, and (2) freezing the backbone network also has some positive effects.

Moreover, we still conduct experiments with the general triplet-based loss function (Eqn (\ref{eq:gtblf_})), setting $w_{ij} = 1$ for all triplets. We empirically set the margin $m = 0.1$. It achieves a little worse results compared to the pair-based loss function. Therefore, we focus on pair-based loss function to demonstrate the effectiveness of samples mining in the following.

\subsubsection{The effectiveness of samples mining}
There are two places  referring to the samples mining. One is the sampler strategy to construct the mini-batch during training. The other is the informative samples mining in the loss function. We test our pair-based baseline with different sampler strategies (random sampler and $PK$ sampler) and the informative samples mining methods (pair distance thresholds and triplet margin).  The results are shown in Table \ref{tab:sample}.

\begin{table}
\caption{The results of samples mining on the CUB-200-2011.}
\label{tab:sample}
  \centering
  \begin{tabular}{c|c|c|c|c|c|c}
    \toprule[2pt]
    Racall@K (\%)  & 1 & 2 & 4 & 8 & 16 & 32 \\ \toprule[1pt]
     Random sampler & 52.0 & 64.0  & 74.7 & 83.1 & 89.8 & 93.9 \\ \hline
     baseline & \multirow{3}{*}{65.5} & \multirow{3}{*}{76.6} & \multirow{3}{*}{84.7} & \multirow{3}{*}{90.8} & \multirow{3}{*}{94.3} & \multirow{3}{*}{97.3} \\
     ($PK$ sampler, & & & & & &  \\
     thresholds) & & & & & & \\ \hline
     + triplet margin & 64.8 & 76.1 & 84.4 & 90.5 & 94.7 & 97.3 \\ \toprule[2pt]
  \end{tabular}
\end{table}
Our baseline method is with $PK$ sampler. For mini-batch construction, the random sampler is adopted to compare with $PK$ sampler. The results in Table \ref{tab:sample} show that random sampler performs much worse than $PK$ sampler, which demonstrates the effectiveness of $PK$ sampler for constructing informative mini-batch. It can reflect the structure relationship of samples, which is stressed by lifted structured loss \cite{ohS2016deep} and N-pair loss \cite{sohn2016improved}.

Moreover, our baseline method (Eqn. (\ref{eq:gpblf_})) implicitly includes the pair distance thresholds for the informative samples mining. Therefore, we add the triplet margin mining (Eqns. (\ref{eq:tripmp} and (\ref{eq:tripmn})) after the pair distance thresholds mining. We empirically set $m_1 = 0$\footnote{Due to the $PK$ sampler to construct mini-batch, for every query sample there are only $K-1$ positive samples while $(P-1)K$ negative samples. We set $m_1 = 0$, without mining the positive samples.}, $m_2 = 0.8$ for pair distance thresholds mining, and $\epsilon = 0.1$ for triplet margin mining. The results show that adding triplet margin mining do not improve the performance. To some extent, it demonstrates the effectiveness of $PK$ sampler combined with pair distance thresholds for informative samples mining.

\subsubsection{The effectiveness of pairs weighting}
\label{sssec:pairw}
There are four aspects about the pair weighting, square distance, $p_{th}$ power weighting, exponential weighting and weight normalization. We conduct experiments based on both pair-based weighting loss (Eqn. (\ref{eq:gpblf_})) and triplet-based weighting loss (Eqn. (\ref{eq:gtblf_})) to evaluate the effectiveness of pairs weighting. For simplification, along with the different parameters in weight computation, we set three kinds of experimental manners.
v0: the baseline architecture, the black lines in Fig. \ref{fig:framework}.  v1: the baseline architecture with weight normalization (Sec. \ref{sssec:wnorm}).  v2: the baseline architecture with weight normalization and multi-level features fusion, adding the red lines in Fig. \ref{fig:framework}.
The results are shown in Table \ref{tab:weight}, from which we can know the following points.

\begin{table*}
\caption{The results of pairs weighting with different hyper-parameters on the CUB-200-2011. We conduct experiments utilizing two kinds of loss functions, pair-based and triplet-based, with two kinds of weighting techniques, $p_{th}$ power weighting and exponential weighting.}
\label{tab:weight}
\centering
\begin{tabular}{l|c|c|c|c|c|c|c|c||c|c|c|c|c|c|c}
\toprule[2pt]
\multicolumn{2}{c|}{} & \multicolumn{7}{c||}{Pair-based loss (Eqn. (\ref{eq:gpblf_}))} & \multicolumn{7}{c}{Triplet-based loss (Eqn. (\ref{eq:gtblf_}))} \\  \hline %\cline{2-9}
  \multicolumn{3}{c|} {Racall@K (\%)}  &  1 & 2 & 4 & 8 & 16 & 32 & & 1 & 2 & 4 & 8 & 16 & 32 \\ \hline
  \multicolumn{3}{c|} {Square} & 65.2  & 76.5 & 84.8 & 90.8 & 94.8 & 97.4 &  & 63.9 & 76.2 & 84.8 & 90.8 & 95.0 & 97.4 \\ \hline
  \multirow{11}{*}{\begin{sideways}{\minitab[l]{Power weighting \\ Eqns. (\ref{eq:powerp}, \ref{eq:powern}, \ref{eq:powert})}}\end{sideways}}
  & v0 & \multirow{3}{*}{\minitab[c]{$p,q=0,0$ }} & 63.4 & 74.2 & 82.9 & 89.2 & 93.9 & 97.1 &  & 60.9 & 73.4 & 82.6 & 90.0 & 94.6 & 97.1\\
  & v1 & & 65.5 & 76.6 & 84.7 & 90.8 & 94.3 & 97.3 & $p=0$ & 64.3 & 75.8 & 84.5 & 90.4 & 95.0 & 97.3 \\
  & v2 & & 68.1 & 79.0 & 86.7 & 92.1 & 95.5 & 97.8 &  & 66.5 & 77.6 & 86.1 & 91.4 & 95.3  & 97.7 \\ \cline{2-16}
  & v0 &\multirow{3}{*}{\minitab[c]{$p,q=0,1$ }} & 56.9 & 68.4 & 78.1 & 86.1 & 91.7 & 95.2  &  & 61.1 & 72.9 & 82.3 & 89.3 & 94.4  & 97.0 \\
  & v1 & & 64.1 & 75.5 & 84.2 & 90.4 & 94.4 & 97.2 & $p=1$ & 63.6 & 75.3 & 84.2 & 90.9 & 94.7  & 97.3 \\
  & v2 & & \textbf{68.6} & 79.3 & 86.7 & 92.0 & 95.5 & 97.8 &  & 67.1 & 78.1 & 85.8 & 91.7 & 95.5 & 97.8\\ \cline{2-16}
  & v2 & $p,q=0,0.5$ & 68.4 & 78.3 & 86.4 & 92.3 & 95.9 & 97.8 & $p=0.5$ & 67.0 & 77.7 & 85.8 & 91.8 & 95.3 & 97.5\\
  & v2 & $p,q=0.5,0.5$ & 68.0 & 78.4 & 86.4 & 92.0 & 95.6 & 97.8 & $p=2$ & 67.8 & 79.1 & 86.4 & 91.8 & 95.5 & 97.5\\
  & v2 & $p,q=0,2$ & 68.1 & 78.4 & 86.3 & 91.6 & 95.4 & 97.8 & $p=3$ & 67.7 & 78.4 & 86.2 & 91.8 & 95.4 & 97.7\\
  & v2 & $p,q=0,4$ & 68.1 & 79.0 & 86.9 & 92.0 & 95.4 & 97.6 & $p=4$ & 67.8 & 78.0 & 86.1 & 91.4 & 95.2 & 97.5\\
  & v2 & $p,q=0,5$ & 67.7 & 78.5 & 86.7 & 91.6 & 95.7 & 97.5 & $p=5$ & \textbf{68.4} & 78.8 & 85.8 & 91.3 & 95.1 & 97.5\\
  \toprule[1pt] \toprule[1pt]
  \multirow{10}{*}{\begin{sideways}{\minitab[l]{Exponential weighting \\ Eqns. (\ref{eq:exponentialp}, \ref{eq:exponentialn}, \ref{eq:exponentialt})}}\end{sideways}}
  & v0 & \multirow{3}{*}{\minitab[c]{$\alpha,\beta=0,1$ }}  & 63.3 & 75.1 & 83.8 & 90.3 & 94.4 & 97.0 &  & 61.2 & 73.5 & 82.4 & 89.2 &  93.6 & 96.7\\
  & v1 &   & 64.8 & 75.8 & 84.6 & 91.0 & 95.0 & 97.3 & $\alpha=1$ & 63.4 & 74.2 & 83.5 & 90.0 & 94.1  & 96.7\\
  & v2 &   & 67.6 & 78.7 & 86.8 & 92.3 & 95.4 & 97.8 &  & 66.5 & 77.2 & 85.5 & 91.9 &  95.6 & 97.8\\ \cline{2-16}
  & v2 &  $\alpha,\beta=0,2$ & \textbf{68.6} & 78.8 & 86.5 & 92.1 & 95.9 & 97.8 & $\alpha=2$ & 67.5 & 78.4 & 86.5 & 92.1 & 95.6  & 97.6\\
  & v2 &  $\alpha,\beta=1,2$ & 66.7 & 77.5 & 85.9 & 91.8 & 95.2 &97.6  & $\alpha=3$ & 66.4 & 77.9 & 86.2 & 91.7 & 95.4  & 97.6\\
  & v2 &  $\alpha,\beta=2,2$ & 67.3 & 77.8 & 85.4 & 91.7 & 95.6 & 97.8 & $\alpha=5$ & 66.8 & 77.5 & 85.4 & 91.6 & 9.3  & 97.6\\
  & v2 &  $\alpha,\beta=0,5$ & 68.0 & 78.3 & 86.3 & 92.0 & 95.7 & 97.8 & $\alpha=10$ & 67.2 & 77.7 & 86.3 & 91.6 & 95.4  & 97.4\\
  & v2 &  $\alpha,\beta=0,10$ & 68.1 & 78.4 & 86.3 & 92.0 & 95.5 & 97.6 & $\alpha=20$ & 67.7 & 78.7 & 86.2 & 91.4 & 95.3  & 97.6\\
  & v2 &  $\alpha,\beta=0,20$ & 68.2 & 78.8 & 86.2 & 91.8 & 95.4 & 97.7 & $\alpha=40$ & \textbf{68.5} & 78.7 & 86.8 & 92.1 & 95.3  & 97.7\\
  & v2 &  $\alpha,\beta=0,40$ & 68.3 & 78.7 & 86.4 & 91.8 & 95.2 & 97.4 & $\alpha=60$ & 67.8 & 78.7 & 86.9 & 92.0 & 95.6  & 97.7\\

\toprule[2pt]
\end{tabular} \\
\footnotesize{v0: The baseline architecture, the black lines in Fig. \ref{fig:framework}.  v1: The baseline architecture with weight normalization (Sec. \ref{sssec:wnorm}).  v2: The baseline architecture with weight normalization and multi-level features fusion, adding the red lines in Fig. \ref{fig:framework}. }\\
\end{table*}

\begin{figure}
\centering
\includegraphics[width=90mm]{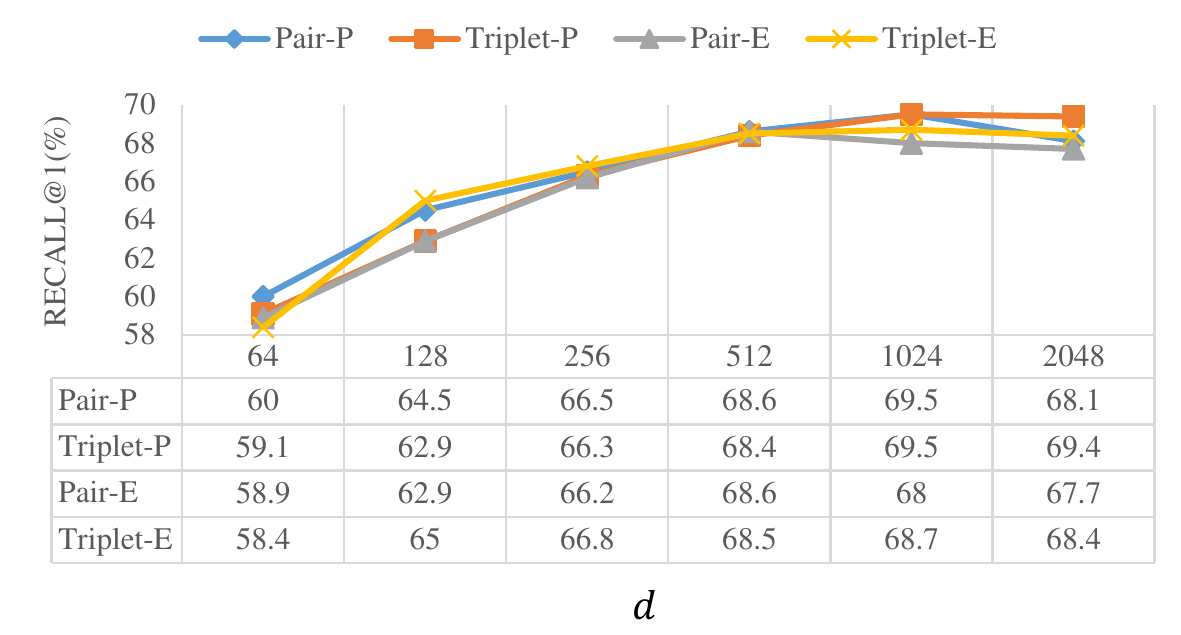}
\caption{The effect of feature dimensions with our proposed general loss functions on CUB-200-2011 dataset.}
\label{fig:dim}
\end{figure}

$\bullet$ \textbf{Square distance}. Compared to v1 with $p,q=0,0$ (our baseline), square distance\footnote{Due to the the large difference between the numbers of positive and negative pairs, we perform the average operation. Therefore, it is compared to v1 with $p,q=0,0$.} (Eqns. (\ref{eq:squarep}) and (\ref{eq:squaret})) achieved comparable results. However, square distance can not combine with our pair weighting methods, because the square operation of distance will destroy the designed weights.

$\bullet$ \textbf{Weight normalization}. v1 performs much better than v0, demonstrating the effectiveness of weight normalization. The weight normalization could take those relative pair distances into account, which could reflect the structure information of training samples in a mini-batch. When $p,q=0,0$, the weight normalization is just the average operation, which is very important due to the large difference between the numbers of positive and negative pairs.

$\bullet$ \textbf{Multi-level feature fusion}. v2 performs much better than v1, demonstrating the effectiveness of multi-level feature fusion, which could obtain  more discriminative features compared to single-level feature.

$\bullet$ \textbf{Pair-based weighting loss vs. triplet-based weighting loss}. Pair-based weighting loss (Eqn. (\ref{eq:gpblf_})) and triplet-based weighting loss (Eqn. (\ref{eq:gtblf_})) can achieve comparable results through parameters finetuning.

$\bullet$ \textbf{$p_{th}$ power weighting vs. exponential weighting}. $p_{th}$ power weighting and exponential weighting also can achieve comparable results through parameters finetuning. In our experiments, under pair-based loss (Eqn. (\ref{eq:gpblf_})) the best results are obtained when $p=0$ for $p_{th}$ power weighting and $\alpha=0$ for exponential weighting. It is similar to the lifted structured loss \cite{ohS2016deep}, which also assign $w_{ij}=1$ for positive pairs, focusing on weighting those informative negative pairs.

\subsubsection{The feature dimensions} To explore the effect of feature dimensions, we conducted experiments with varying fully connected layer dimensions $\{64, 128, 256, 512, 1024, 2048\}$, using our proposed loss functions.
\begin{itemize}
\item \emph{Pair-P} denotes the pair-based loss with $p_{th}$ power weighting, setting $p,q=0,1$.
\item \emph{Triplet-P} denotes triplet-based loss with $p_{th}$ power weighting, setting $p=5$.
\item \emph{Pair-E} denotes the pair-based loss with exponential weighting, setting $\alpha,\beta=0,2$.
\item \emph{Triplet-E} denotes denotes triplet-based loss with exponential weighting, setting $\alpha=40$.
\end{itemize}
 Thanks to the multi-level features fusion, when the output dimensions of fully connected layer is $d$, the dimension of the final features is $3d$. The recall@1 results are shown in Fig. \ref{fig:dim}.

The performance of all four methods are consistently increasing with $d$ in $\{64, 128, 256, 512\}$. \emph{Pair-P}, \emph{Triplet-P}, and \emph{Triplet-E} achieve the best performance when $d = 1024$, while \emph{Pair-E} achieve the best performance when $d = 512$.

\subsubsection{Comparison to multi-similarity loss \cite{wang2019multi}}

\begin{figure}
\centering
\includegraphics[width=80mm]{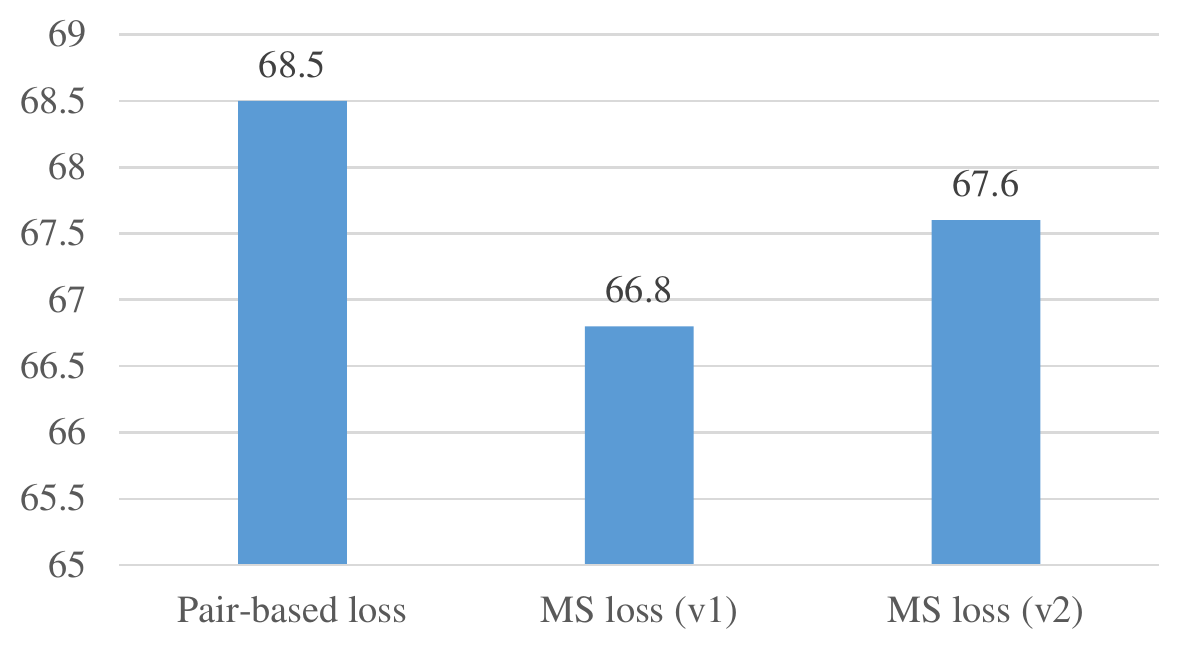}
\caption{The comparison results (Recall@1 (\%)) of our pair-based loss with multi-similarity loss on CUB-200-2011 dataset.}
\label{fig:comp}
\end{figure}

We compare our pair-based weighting loss function with exponential weighting (also weight normalization and multi-level features fusion) to the multi-similarity loss \cite{wang2019multi}, since multi-similarity loss is the most similar one to our method and achieves the state-of-the-art performance. For fair comparison, we adopt the parameters from its original paper, setting $\alpha,\beta = 2,50$ for both loss functions, and $m_1 = 0$, $m_2=1$ for our loss, while $m = 1$ for multi-similarity loss. Moreover, our pair-based weighing loss adopts pair distance thresholds (Eqns. (\ref{eq:smpp}) and (\ref{eq:smpn})) for samples mining, while multi-similarity loss adopts triplet margin mining (Eqns. (\ref{eq:tripmp} and (\ref{eq:tripmn})), setting $\epsilon=0.1$.
In this case, the two loss functions can be mathematically formulated as follows.

Our pair-based loss,
\begin{align}
    \mathcal{L} (\{D_{ij}\}) = \sum_{y_{ij} = 1} w_{ij}^{+} [D_{ij}-m_1]_{+} + \sum_{ y_{ij} = 0} w_{ij}^{-} [m_2 - D_{ij}]_{+}, \label{eq:ourloss}
\end{align}
where $w_{ij}^{+} = \frac{\exp(\alpha(D_{ij}-m_1))}{\sum_{y_{ij} = 1} \exp(\alpha(D_{ij}-m_1))}$, and $w_{ij}^{-} = \frac{\exp(\beta(m_2 - D_{ij}))}{\sum_{y_{ij} = 0} \exp(\beta(m_2 - D_{ij}))}$.

The multi-similarity loss,
\begin{align}
    \mathcal{L} (\{D_{ij}\}) &= \frac{1}{\alpha} \log \Big[1+ \sum_{y_{ij} = 1}\exp \big(\alpha(D_{ij} - m)\big)\Big] \nonumber \\
     &+ \frac{1}{\beta} \log \Big[1+ \sum_{y_{ij} = 0}\exp \big(\beta(m - D_{ij})\big)\Big], \nonumber
\end{align}
which can be uniformly translated into our general pair-based loss formulation as follows,
\begin{align}
    \mathcal{L} (\{D_{ij}\}) = \sum_{y_{ij} = 1} w_{ij}^{+} (D_{ij}-m) + \sum_{ y_{ij} = 0} w_{ij}^{-} (m - D_{ij}), \label{eq:ourloss}
\end{align}
where $w_{ij}^{+} = \frac{\exp(\alpha(D_{ij}-m))}{1+\sum_{y_{ij} = 1} \exp(\alpha(D_{ij}-m))}$, and $w_{ij}^{-} = \frac{\exp(\beta(m - D_{ij}))}{1+\sum_{y_{ij} = 0} \exp(\beta(m - D_{ij}))}$.

From the loss functions, we can find that.
\begin{itemize}
\item The hinge function $[\cdot]_{+}$ in our loss would make all the terms ($D_{ij} - m_1$ and $m_2 - D_{ij}$) in weights be positive, while in multi-similarity loss those terms may be negative.
\item Our loss has two threshold parameters $m_1$ for positive pairs and $m_2$ for negative pairs, while multi-similarity loss only has one $m$ for both positive and negative pairs.
\item Compared to our method, the weights of multi-similarity loss adds 1 in the denominator.
\item In multi-similarity loss $\alpha$ and $\beta$ can not be equal to 0, setting constant weight, while our loss could.
\end{itemize}

We conduct experiments with the following losses, our pair-based loss, the original multi-similarity loss (MS loss (v1)) and the multi-similarity loss without adding 1 (MS loss (v2)). The corresponding results are shown in Fig. \ref{fig:comp}.

(1) MS loss (v2) performs better than MS loss (v1). It demonstrates that adding 1 is not a good operation for deep metric learning. From the perspective of weighting, $1 = \exp(m - m)$ which may introduce sample pairs whose distance is $m$ for both positive and negative pairs, which may increase the training difficulty.

(2) Our pair-based loss outperforms MS loss (v2), which demonstrates the effectiveness of our the samples mining and pairs weighting methods compared to those of multi-similarity loss.

\begin{table*}
\caption{Comparison with the state-of-the-art methods on CUB-200-2011 and Cars196 datasets in terms of Recall@K (\%). The hyper-parameters for weighing are listed in the brackets along with the corresponding methods. The '-' denotes the corresponding results are not reported in the original paper.}
\label{tab:cub_car}
\centering
\begin{tabular}{l|c|c|c|c|c|c|c||c|c|c|c|c|c}
\toprule[2pt]
Method & Publication & \multicolumn{6}{c||}{CUB-200-2011} & \multicolumn{6}{c}{Cars196} \\ \hline
\multicolumn{2}{c|}{Racall@K (\%)}&  1 & 2 & 4 & 8 & 16 & 32 & 1 & 2 & 4 & 8 & 16 & 32 \\ \hline
Lifted structure loss \cite{ohS2016deep} & CVPR16 & 43.6 & 56.6 & 68.6 & 79.6 & - & - & 53.0 & 65.7 & 76.0 & 84.3 & - & -\\
N-pair loss \cite{sohn2016improved} & NeurIPS16 & 51.0& 63.3 & 74.3 & 83.2 & - & - & 71.1 & 79.7 & 86.5 & 91.6 & -& -\\
Spectral clustering \cite{law2017deep} & ICML17 &53.2 & 66.1 & 76.7 & 85.3 & -& -& 73.1  & 82.2 & 89.0 &93.0 & - & - \\
Clustering \cite{oh2017deep} & CVPR17& 48.2 & 61.4 & 71.8 & 81.9 & & & 58.1& 70.6 & 80.3 & 87.8 & & \\
proxy-NCA \cite{movshovitz2017no} & ICCV17 & 49.2 & 61.9 & 67.9 & 72.4 & - & - & 73.2 & 82.4 & 86.4 & 87.8 &- & -\\
Deeply cascaded \cite{yuan2017hard} & ICCV17 & 53.6 & 65.7 & 77.0 & 85.6 & 91.5 & 95.5 & 73.7 & 83.2 & 89.5 & 93.8 & 96.7 & 98.4 \\
Margin \cite{wu2017sampling} & ICCV17 &63.6 & 74.4 & 83.1 & 90.0 & 94.2 & - & 79.6 & 86.5 & 91.9 & 95.1 & 97.3 & - \\
Deep adversarial \cite{duan2018deep}  & CVPR18 & 52.7 & 65.4 & 75.5 & 84.3 & - & - & 75.1 & 83.8 & 89.7 & 93.5 & - & - \\
ABE \cite{kim2018attention} & ECCV18 & 60.6 & 71.5 & 79.8 & 87.4 & - & - & 85.2  & 90.5 & 94.0 & 96.1 & - & - \\
Hierarchical triplet loss \cite{ge2018deep} & ECCV18 & 57.1& 68.8 & 78.7 & 86.5 & 92.5 & 95.5 & 81.4 & 88.0 & 92.7 & 95.7 & 97.4 & 99.0 \\
ABIER \cite{opitz2018deep} & TPAMI18 & 57.5 & 68.7 & 78.3 & 86.2 & 91.9 & 95.5 & 82.0 & 89.0 & 93.2 & 96.1 & 97.8 & 98.7 \\
Hardness-aware \cite{zheng2019hardness} & CVPR19 & 53.7 & 65.7 & 76.7 & 85.7 & - & - & 79.1 & 87.1 & 92.1 & 95.5 & - & - \\
Ranked list loss \cite{wang2019ranked} & CVPR19 & 61.3 & 72.7 & 82.7 & 89.4 & - & - & 82.1 & 89.3 & 93.7 & 96.7 & - & - \\
Multi-similarity loss \cite{wang2019multi} & CVPR19 & 65.7& 77.0 & 86.3 & 91.2 & 95.0 & 97.3 & 84.1 & 90.4 & 94.0 & 96.5 & 98.0 & 98.9 \\ \hline \hline
Pair-P ($p,q,d=0,1,1024$) & ours & \textbf{69.5} & \textbf{79.9} & \textbf{87.2} & 91.9 & \textbf{95.8} & \textbf{97.9} & 86.3 & 91.8 & \textbf{95.1} & \textbf{97.4} &98.6 & \textbf{99.3} \\
Trilet-P ($p,d=5,1024$) & ours & \textbf{69.5} & \textbf{79.9} & 86.8 & 92.0 & 95.6 & 97.7 & 84.8 & 91.1 & 94.1 & 96.5 & 98.0 & 98.9 \\
Pair-E ($\alpha,\beta,d=0,2,1024$)  & ours & 68.0 & 78.6 & 86.7 & \textbf{92.1} & 95.6 & \textbf{97.9} & \textbf{86.5} & \textbf{92.1} & \textbf{95.1} & \textbf{97.4} & \textbf{98.7} & \textbf{99.3}\\
Trilet-E ($\alpha,d=40,1024$) & ours & 68.7 & 78.9 & 86.4 & \textbf{92.1} & 95.5 & 97.8 & 84.9 & 91.0 & 94.3 & 96.4 & 98.0 & 98.9 \\
\toprule[2pt]
\end{tabular}
\end{table*}

\subsection{Comparison with state-of-the-art}
We further compare the performance of our proposed general pair-based weighting loss with the state-of-the-art methods on image retrieval tasks. We empirically set $m_1 = 0$, $m_2 = 0.8$, $m = 0.1$ for CUB-200-2011 and Cars196 datasets, while $m_1 = 0$, $m_2 = 1.0$, $m = 0.2$ for SOP dataset.
The other hyper-parameters for weighing are listed in the brackets along with the corresponding methods in Tables \ref{tab:cub_car} and \ref{tab:sop}.
We can find the following observations.

\begin{table}
%\scriptsize
\caption{Comparison with the state-of-the-art methods on SOP dataset in terms of Recall@K (\%). The hyper-parameters for weighing are listed in the brackets along with the corresponding methods. The '-' denotes the corresponding results are not reported in the original paper.}
\label{tab:sop}
\centering
\begin{tabular}{l|c|c|c|c}
\toprule[2pt]
Method & Publication & \multicolumn{3}{c}{SOP} \\ \hline
\multicolumn{2}{c|}{Racall@K (\%)}&  1 & 10 & 100  \\ \hline
 Lifted structure loss \cite{ohS2016deep} & CVPR16 & 62.5 & 80.8 & 91.9  \\
 N-pair loss \cite{sohn2016improved} & NeurIPS16 & 67.7 & 83.8 & 93.0  \\
 Clustering \cite{oh2017deep} & CVPR17& 67.0 & 83.7& 93.2  \\
 proxy-NCA \cite{movshovitz2017no} & ICCV17 & 73.7 &- &-   \\
 Margin \cite{wu2017sampling} & ICCV17 & 72.7 & 86.2 & 93.8  \\
 Deeply cascaded \cite{yuan2017hard} & ICCV17 & 69.5 & 84.4 & 92.8  \\
 Deep adversarial \cite{duan2018deep}  & CVPR18 & 68.4 & 83.5 & 92.3  \\
 Hierarchical triplet loss \cite{ge2018deep} & ECCV18 & 74.8 & 88.3 & 94.8  \\
 ABE \cite{kim2018attention} & ECCV18 & 76.3 & 88.4 & 94.8  \\
 ABIER \cite{opitz2018deep} & TPAMI18 & 74.2 & 86.9 & 94.0  \\
 Hardness-aware \cite{zheng2019hardness} & CVPR19 & 68.7 & 83.2 & 92.4  \\
 Ranked list loss \cite{wang2019ranked} & CVPR19 & 79.8 & 91.3 & 96.3  \\
 Multi-similarity loss \cite{wang2019multi} & CVPR19 & 78.2 & 90.5 & 96.0  \\ \hline \hline
 Pair-P ($p,q,d=1,5,1024$)  & ours & 80.8 & 91.9 & 96.9 \\
 Triplet-P ($p,d=5,1024$) & ours & 79.8 & 91.5 & 96.5 \\
 Pair-E ($\alpha,\beta,d=2,10,1024$) & ours &\textbf{80.9} & \textbf{92.1} & \textbf{96.9} \\
 Triplet-E ($\alpha,d=40,1024$) & ours & 80.3 & 91.6 & 96.7 \\
\toprule[2pt]
\end{tabular}
\end{table}

\begin{itemize}
\item Our proposed general weighting losses can achieve the best performance against all the compared methods, demonstrating the effectiveness of our samples mining and pais weighting methods in our general weighting loss formulations.
\item Our proposed general weighting losses outperform not only those existing pair-based losses (lifted structure loss \cite{ohS2016deep}, N-pair loss \cite{sohn2016improved}, margin \cite{wu2017sampling}, hierarchical triplet loss \cite{ge2018deep}, ranked list loss \cite{wang2019ranked}, multi-similarity loss \cite{wang2019multi}), but also clustering-based methods (clustering \cite{oh2017deep}, spectral clustering \cite{law2017deep}, proxy-NCA \cite{movshovitz2017no}), adversarial techniques (deep adversarial \cite{duan2018deep}, hardness-aware \cite{zheng2019hardness}) and even ensemble approaches (ABE \cite{kim2018attention}, ABIER \cite{opitz2018deep}).
\item On the fine-grained datasets, like CUB-200-2011 and Cars196, our methods achieve new state-of-the-art performance much better than the previous state-of-the-art. While on the large-scale dataset with enormous categories, like SOP, our methods can achieve a little better (or comparable) performance compared to the previous state-of-the-art.
\end{itemize}

\begin{figure*}
\centering
\includegraphics[width=170mm]{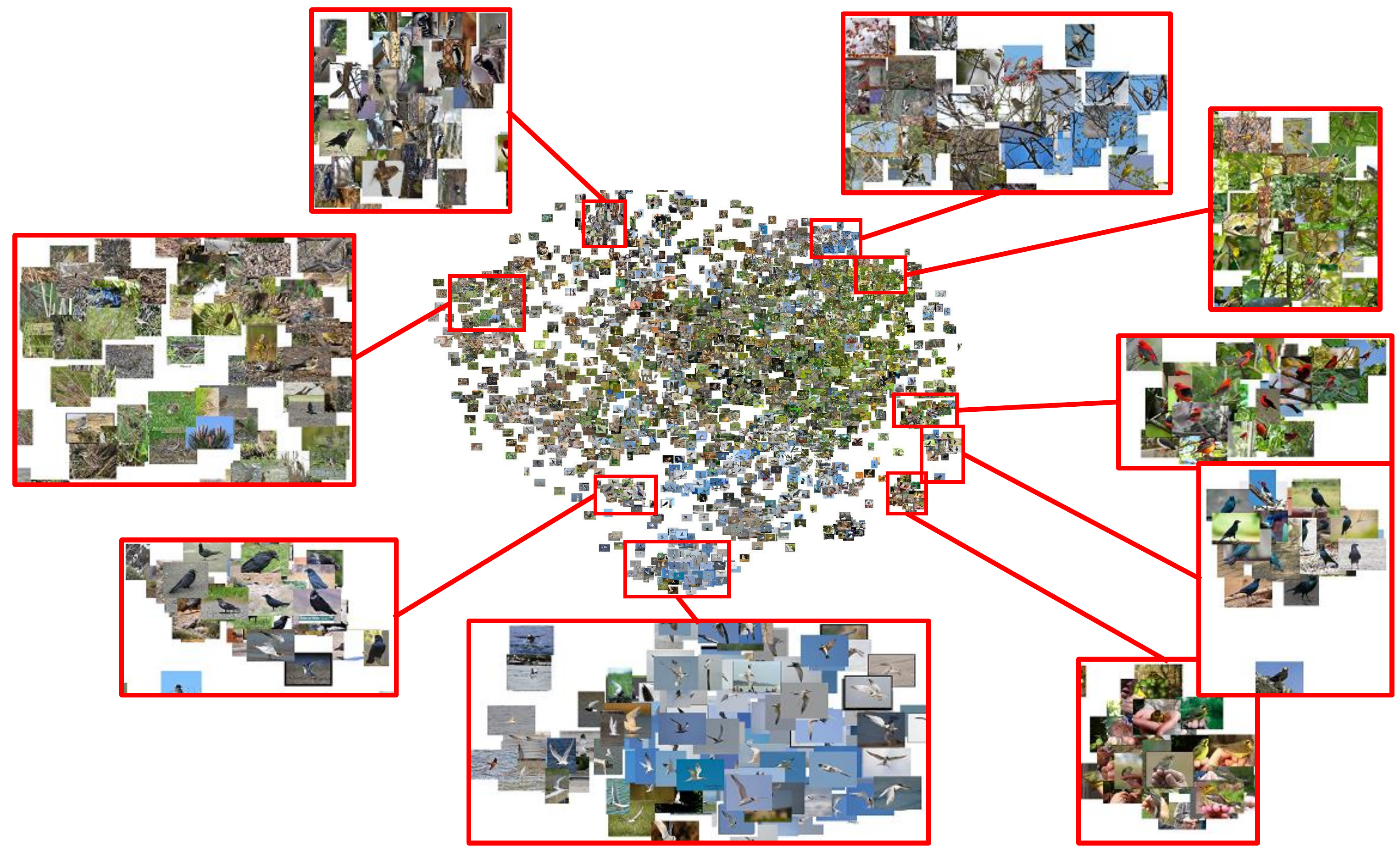}
\caption{Barnes-Hut t-SNE visualization \cite{van2014accelerating} of our embeddings on the test split (class 101 to 200; 5,924 images) of CUB-200-2011. Best viewed on a monitor when zoomed in.}
\label{fig:vis_cub}
\end{figure*}

\begin{figure*}
\centering
\includegraphics[width=170mm]{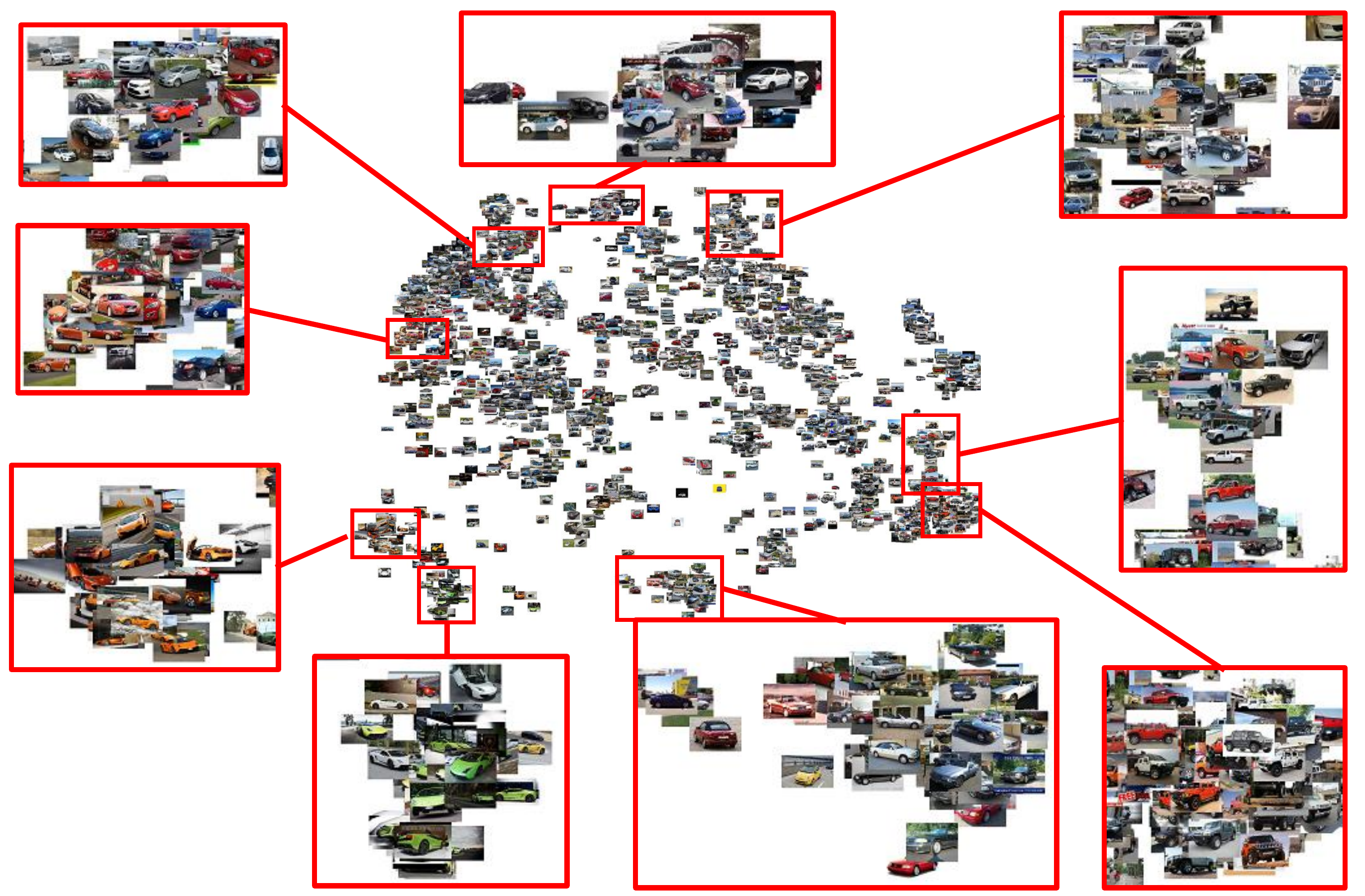}
\caption{Barnes-Hut t-SNE visualization \cite{van2014accelerating} of our embeddings on the test split (class 99 to 196; 8,131 images) of Cars196. Best viewed on a monitor when zoomed in.}
\label{fig:vis_car}
\end{figure*}

\noindent \textbf{Qualitative Results:} Figs. \ref{fig:vis_cub} and \ref{fig:vis_car} respectively show the Barnes-Hut t-SNE visualization \cite{van2014accelerating} of the learned embeddings of CUB-200-2011 and Cars196 by our proposed pair-based weighting loss. Since the two datasets are fine-grained datasets, the visual differences between two classes tend to be very subtle, which makes it difficult for human beings to distinguish. We observe that despite the subtle inter-class differences and large intra-class variations, such as backgrounds, viewpoints, illuminations and poses, our methods can still be able to cluster those images from the same class.

\section{Conclusions}
In this paper, based on the existing pair-based losses, we present a general pair-based weighting loss formulation for deep metric learning to discriminatively learning the semantic embeddings. It casts the pair-based loss into two aspects, samples mining and pairs weighting. Samples mining aims at selecting the informative pairs in order to exploit the structure relationship among a mini-batch. Pairs weighting aims at assigning different weights for different pairs to make them contribute differently.
We reviewed those existing pair-based losses inline with our general weighting loss formulation. Some possible samples mining and pairs weighting techniques are detailedly explored, and the potential combinations of them are also considered.
Extensive experiments on three image retrieval benchmarks with new state-of-the-art performance demonstrate the effectiveness of our general pair-based weighting loss for deep metric learning.

\section*{Acknowledgment}

This work was supported by the National Natural Science Foundation of China (61671125, 61201271, 61301269), and the State Key Laboratory of Synthetical Automation for Process Industries (NO. PAL-N201401).

\bibliographystyle{IEEEtrans}
\bibliography{egbib}

\end{document}